\documentclass[a4paper,fleqn]{cas-sc}

\usepackage[numbers]{natbib}
\usepackage{hyperref}
\usepackage{url}
\usepackage{graphicx}    
\usepackage{subcaption} 
\usepackage{booktabs}

\usepackage{algorithm}
\usepackage{algpseudocode}
\def\tsc#1{\csdef{#1}{\textsc{\lowercase{#1}}\xspace}}
\tsc{WGM}
\tsc{QE}

\begin{document}
\let\WriteBookmarks\relax
\def\floatpagepagefraction{1}
\def\textpagefraction{.001}

\shorttitle{Partner-Aware Hierarchical Skill Discovery for Robust Human-AI Collaboration}    

\shortauthors{}  

\title [mode = title]{Partner-Aware Hierarchical Skill Discovery for Robust Human-AI Collaboration}  



%

\author[1]{Adnan Ahmad}

\cormark[1]


\ead{adnana0111@gmail.com}


\credit{Conceptualization, Formal analysis, Investigation, Data curation, Writing - original draft, Writing - review \& editing, Visualization. }

\affiliation[1]{organization={Schoold of Information Technology, Deakin University},
            addressline={75 Pidgons Rd}, 
            city={Waurn Ponds},
            postcode={3216}, 
            state={VIC},
            country={Australia}}

\author[1]{Bahareh Nakisa}




\credit{Conceptualization, Methodology, Writing – review \& editing, Supervision, Project administration, Funding acquisition}
\author[2]{Mohammad Naim Rastgoo}
\affiliation[2]{organization={Faculty of Information Technology, Monash University},
            addressline={Wellington Rd}, 
            city={Clayton},
            postcode={3800}, 
            state={VIC},
            country={Australia}}
\credit{Validation, Investigation}
\cortext[1]{Corresponding author}



\begin{abstract}
Multi-agent collaboration, especially in human-AI teaming, requires agents that can adapt to novel partners with diverse and dynamic behaviors. Conventional Deep Hierarchical Reinforcement Learning (DHRL) methods focus on agent-centric rewards and overlook partner behavior, leading to shortcut learning, where skills exploit spurious information instead of adapting to partners’ dynamic behaviors. This limitation undermines agents' ability to adapt and coordinate effectively with novel partners. We introduce Partner-Aware Skill Discovery (PASD), a DHRL framework that learns skills conditioned on partner behavior. PASD introduces a contrastive intrinsic reward to capture patterns emerging from partner interactions, aligning skill representations across similar partners while maintaining discriminability across diverse strategies. By structuring the skill space based on partner interactions, this approach mitigates shortcut learning and promotes behavioral consistency, enabling robust and adaptive coordination. We extensively evaluate PASD in the Overcooked-AI benchmark with a diverse population of partners characterized by varying skill levels and play styles. We further evaluate the approach with human proxy models trained from human–human gameplay trajectories. PASD consistently outperforms existing population-based and hierarchical baselines, demonstrating transferable skill learning that generalizes across a wide range of partner behaviors. Analysis of learned skill representations shows that PASD adapts effectively to diverse partner behaviors, highlighting its robustness in human-AI collaboration.
\end{abstract}




\maketitle

\section{Introduction}\label{}
Developing intelligent agents that can coordinate effectively with humans and other novel partners has long been a central challenge in multi-agent reinforcement learning (MARL) \citep{Klein2004TenCF, Alami2006TowardHR, bard2020hanabi}. Unlike adversarial settings \citep{ye2020towards}, where success is measured by outperforming an opponent, collaboration is far more challenging as it requires adapting to novel partners with diverse and often unpredictable behaviors \citep{hu2020other}. Early approaches relied on behavior cloning (BC) from human–human interaction data, but these methods are costly, time-consuming, and struggle to capture the diversity of real-world behaviors \citep{carroll2019utility}. \textcolor{black}{Furthermore, even in fully observable environments, partner behaviors are not fully predictable from a single state. Differences in timing, hesitation, movement rhythms, and style preferences unfold over sequences of actions, requiring temporal modeling to capture and adapt to these patterns. Conditioning on the current state is insufficient to avoid behavioral interference or forgetting of previously learned coordination strategies.} More recently, hierarchical reinforcement learning (HRL) has advanced collaboration by decomposing complex tasks into reusable skills, enabling more structured exploration and improved coordination \citep{eysenbach2019diversity, loo2023hierarchical}. HRL provides a framework for structuring agent behavior through temporally extended actions, or 'skills', which can capture reusable patterns of behavior. By learning a set of diverse and reusable skills, agents can explore more efficiently and adapt their behavior in complex environments. 

However, standard skill discovery methods remain largely agent-centric, optimizing for diversity without accounting for partner influence. Consequently, the learned skills may support individual performance but are insufficient for robust coordination with diverse partners. We argue that this limitation arises from the agent-centric nature of reward optimization. Existing approaches maximise expected returns from the agent’s perspective, often ignoring the influence of the partner on cooperative dynamics. This misalignment leads to \emph{shortcut learning} \citep{Wei2023OnlinePL}, where agents exploit spurious correlations in the environment rather than capturing information relevant for partner interactions. As a result, agents develop behaviors that prioritize their own action diversity but fail to generalize coordination across novel partners.

Prior skill discovery methods in HRL often maximize mutual information (MI) between skills and observational states as a proxy for behavioral diversity \cite{Gregor2016VariationalIC, eysenbach2019diversity}. While this encourages the agent to develop distinguishable behaviors, the objective is bounded by the entropy of skills and does not ensure sensitivity to partner behavior. As a result, these approaches often learn simple and static skills with limited adaptability, leading to poor state coverage and weak coordination, as highlighted in recent studies \citep{campos2020explore, jiang2022unsupervised}. Moreover, tractable variational estimators of MI \citep{eysenbach2019diversity}, typically implemented with neural networks optimized via cross-entropy or related objectives, are prone to shortcut learning . In practice, they often capture spurious correlations in state features rather than the interaction patterns relevant for effective collaboration \cite{Wei2023OnlinePL}.

We introduce Partner-Aware Skill Discovery (PASD), an HRL approach for learning skills that adapt to diverse collaborator behaviors. PASD maximizes a variational lower bound on MI between skills and sub-trajectories, encouraging representations that are consistent and reproducible across partner interactions. This is achieved via a contrastive objective that ensures skill representations are discriminative across heterogeneous partners while remaining consistent for partners with similar behaviors. By capturing patterns shaped by partner behavior rather than agent-centric correlations, PASD mitigates shortcut learning and produces skills that generalize across partners, supporting effective partner-adaptive coordination.

In summary, our work makes the following key contributions:
\begin{itemize}
    \item  We introduce PASD, a DHRL framework that enables robust human-AI coordination through skill representations conditioned on partner behavior.
    \item A novel contrastive intrinsic reward is proposed and incorporated into PASD to encourages consistency in skill representations across similar partners by capturing shared patterns from parallel rollouts while maintaining discriminability across diverse partner behaviors. This intrinsic reward, derived from contrastive learning, encourages behavioral diversity across diverse partners, mitigating shortcut learning caused by spurious state information, and directly leveraging partner-relevant information from the observational space.
    \item We extensively evaluate PASD in the Overcooked-AI environment by partnering the agent with a diverse population pool with varying skill levels and play styles. We also evaluate the proposed approach with human proxy models trained from human–human gameplay trajectories, demonstrating that PASD learns transferable skills that generalize effectively across a wide range of partners, enabling robust and adaptive human-AI coordination.
\end{itemize}
\section{Related Work}
Recent work has explored building agents that can coordinate with human partners \cite{carroll2019utility, Hao2024BCRDRLBA}. Carroll et al.~\cite{carroll2019utility} introduced the \emph{Overcooked-AI} environment and trained PPO agents with human proxy models from human gameplay. While improving robustness, this requires extensive and costly human data. Hao et al.~\cite{Hao2024BCRDRLBA} introduce intrinsic rewards to encourage agents to explore states that yield sparse rewards when coordinating with human proxy models. Strouse et al.~\cite{strouse2021collaborating} propose Fictitious Co-Play (FCP), generating a pool of self-play policies and past versions to train adaptive agents without human data. Some works further improve partner heterogeneity using entropy-based objectives during training \citep{pmlr-v139-lupu21a, garnelo2021pick, zhao2023maximum, loo2023hierarchical}. Hidden-utility Self-Play HSP \citep{yu2023learning} extends FCP by modeling human biases as hidden reward functions, generating a diverse policies to train adaptive agents that can cooperate with unseen humans with preferences deviating from environment rewards. \textcolor{black}{Jha et al. \cite{jha2025crossenvironment} propose Cross-Environment Cooperation (CEC), which trains agents across a distribution of environments to acquire general cooperative skills, enabling zero-shot coordination with novel partners. While effective for generalization across tasks, CEC does not explicitly model partner-adaptive skill discovery within a single environment, which is the focus of our method.} While these methods focus on learning a single-level agent policy, effective human-AI coordination requires reasoning over temporally extended behaviors and adapting to partners with diverse and evolving strategies.

HRL provides a framework for reasoning over temporally extended behaviors, making it well-suited for multi-agent and human-AI coordination. By learning policies at multiple temporal levels \citep{SUTTON1999181, flet2019promise, pateria2021hierarchical}, HRL captures high-level strategic planning and low-level execution. Classical approaches such as options \citep{bacon2017option, eysenbach2019diversity} and feudal learning \citep{vezhnevets2017feudal} illustrate temporal hierarchy benefits, extended to cooperative multi-agent settings in recent work \citep{loo2023hierarchical, NEURIPS2023_c276c330}. Methods like DIAYN \citep{eysenbach2019diversity} encourage diverse behaviors using intrinsic rewards maximizing MI between skills and states/actions. However, these agent-centric approaches are prone to shortcut learning, capturing spurious patterns instead of partner-relevant behaviors. Hierarchical Population Training (HIPT) \citep{loo2023hierarchical} adapts HRL to human-AI coordination by shaping the high-level policy via influence-based intrinsic rewards but trains the low-level policy only on extrinsic rewards, risking skill collapse. Our approach introduces a novel intrinsic reward to mitigate shortcut learning, ensuring behavioral consistency across similar partners while remaining discriminative to diverse strategies, supporting adaptive human-AI coordination

\section{Preliminaries}
We consider a multi-agent setting in which two agents collaborate to complete shared tasks, with the framework naturally extending to settings involving more agents. One agent is controlled by a learning policy $\pi_\theta(\cdot \mid s_t)$, while the other is governed by a partner policy $\pi^p(\cdot \mid s_t)$, sampled uniformly from a population of pretrained partners $\mathcal{D}_p$ at the start of each episode. The objective is to train $\pi_\theta(\cdot \mid s_t)$ to achieve high returns when paired with novel partners drawn from a separate evaluation distribution $\mathcal{D}'_p$. The environment is modeled as a two-player Markov game $\mathcal{M} = (\mathcal{S}, \mathcal{A}, \mathcal{A}^p, \mathcal{P}, r, \gamma, \rho_0)$, where $\mathcal{S}$ is the state space, $\mathcal{A}$ and $\mathcal{A}^p$ are the action spaces of the learning agent and the partner, $\mathcal{P}$ is the transition kernel, $r: \mathcal{S} \times \mathcal{A} \times \mathcal{A}^p \to \mathbb{R}$ is the shared team reward, $\gamma \in (0,1)$ is the discount factor, and $\rho_0$ is the initial state distribution. At each timestep $t$, the learning agent selects an action $a_t \sim \pi_\theta(\cdot \mid s_t)$, while the partner executes $a^p_t \sim \pi^p(\cdot \mid s_t)$, and the next state is drawn from $s_{t+1} \sim \mathcal{P}(s_{t+1} \mid s_t, a_t, a^p_t)$.

From the perspective of the learning agent, the effective dynamics marginalize over both the partner’s stochasticity and the population distribution:
\[
\mathcal{P}_{\mathcal{D}_p}(s' \mid s, a) =
\mathbb{E}_{\pi^p \sim \mathcal{D}_p}\,\mathbb{E}_{a^p \sim \pi^p(\cdot \mid s)}\big[\mathcal{P}(s' \mid s, a, a^p)\big].
\]
This formulation emphasizes that the agent must learn a policy that is robust to variations in partner behavior while maximizing expected returns over the population of collaborators. For a given partner $\pi^p$, the return is

\begin{equation}
J(\pi_\theta \mid \pi^p) = \mathbb{E}\!\left[\sum_{t=0}^\infty \gamma^t \, r(s_t, a_t, a^p_t)\right].    
\end{equation}
To encourage robustness to novel partners, the learning objective is the expected return over the partner population:
\begin{equation}
J(\pi_\theta) = \mathbb{E}_{\pi^p \sim \mathcal{D}_p}\big[ J(\pi_\theta \mid \pi^p)\big].
\end{equation}

\paragraph{Hierarchical reinforcement learning:}
In collaborative multi-agent environments, effective coordination requires reasoning over temporally extended behaviors and adapting to partners with diverse and dynamically changing strategies. To capture these aspects, we model the learning agent using a hierarchical policy inspired by the options framework \cite{SUTTON1999181}. Formally, we consider DHRL setup with a high-level manager $\pi_{hi}(z \mid s)$ and a low-level controller $\pi_{lo}(a \mid s, z)$. The high-level manager selects latent skills $z \in \mathcal{Z}$, which guide temporally extended behaviors executed by the low-level controller. Each skill $z_k$ is executed over a segment from $t_k$ to $t_{k+1}-1$, producing cumulative segment reward:
\begin{equation}
\mathcal{R}^Z(s_{t_k}, z_k) = \sum_{t=t_k}^{t_{k+1}-1} \gamma^{t-t_k} r(s_t, a_t, a^p_t), 
\quad z_k \sim \pi_{hi}(\cdot \mid s_{t_k}), \; a_t \sim \pi_{lo}(\cdot \mid s_t, z_k),
\end{equation}
where $a^p_t$ denotes the partner's action at time $t$ and $\gamma \in [0,1]$ is a discount factor. The high-level objective is the expected return over all skill segments:
\begin{equation} \label{eq:J_hi}
J_{hi}(\pi_{hi}, \pi_{lo} \mid \pi^p) = \mathbb{E}\Bigg[\sum_{k=0}^{K-1} \gamma^{t_k} \mathcal{R}^Z(s_{t_k}, z_k) \Bigg].
\end{equation}
Skill execution is controlled by a stochastic termination function $\beta(z, s)$, which determines whether the current skill continues or a new skill should be selected. This allows the manager to adaptively update skills at irregular intervals $T_{hi}$ based on the evolving collaborative context.

Within each skill segment, the low-level controller $\pi_{lo}(a \mid s, z)$ outputs primitive actions conditioned on the current state and the active skill. The low-level policy is trained to reliably realize the intended skill, producing sequences of actions that induce state transitions $s \to s'$ through the environment dynamics. Formally, the low-level objective can be expressed as:
\begin{equation}\label{eq:J_lo}
J_{lo}(\pi_{lo} \mid z, \pi^p) = \mathbb{E}\Bigg[\sum_{t=t_k}^{t_{k+1}-1} \gamma^{t-t_k} r(s_t, a_t, a^p_t)\Bigg],
\end{equation}
where the expectation is conditioned on the currently active skill $z_k$. The high-level manager, low-level controller, and termination function together define the joint hierarchical policy $\pi = (\pi_{hi}, \pi_{lo}, \beta),$ which is optimized end-to-end via proximal policy optimization (PPO) algorithm \citep{schulman2017proximal} to maximize both high-level and low-level objectives. 

\section{Method}

\subsection{Motivation}

While the high-level and low-level objectives in \ref{eq:J_hi} and \ref{eq:J_lo} focus on maximizing extrinsic team rewards, optimizing only for these objectives often leads to skill collapse, where all skills converge to similar behaviors. Each skill independently maximizes cumulative reward without explicit constraints promoting discriminability, which can result in a single skill dominating entire episodes. Under such conditions, the termination function $\beta(z,s)$ cannot effectively differentiate among skills, and the hierarchical policy loses expressive power. Prior works \cite{eysenbach2019diversity, Gregor2016VariationalIC} have addressed skill collapse by introducing intrinsic objectives that maximize the mutual information (MI) between skills and states, $I(Z;S) = H(Z) - H(Z \mid S)$, thereby encouraging skills to induce distinguishable state distributions. These approaches typically separate skill discovery and high-level optimization into two phases: skills are first discovered by mapping them to diverse states, and then the high-level policy is optimized with extrinsic reward on downstream tasks.

However, in collaborative multi-agent environments, the next state is determined jointly by the agent and the partner policy, $s_{t+1} \sim \mathcal{P}(s_{t+1} \mid s_t, a_t, a^p_t),$ so the state distribution induced by a skill $z$ depends not only on the agent policy $\pi$ but also on the partner policy $\pi^p$.

\textbf{Assumption 1:} The skill space is lower-dimensional than the joint state space, i.e., $H(Z) < H(S), \quad s \sim \rho^{\pi, \pi^p}(s).$ This reflects the fact that each skill typically corresponds to a sub-trajectory of the high-dimensional state space, allowing distinct behaviors to be captured as separate skills.

Under Assumption 1, maximizing $I(Z;S)$ is limited by the low dimensionality of the skill space. Since $H(Z)$ is fixed, the agent can achieve the maximum MI even by producing only minor, agent-centric variations that capture very little meaningful information about the partner-conditioned dynamics. For instance, consider two policies $\pi_1$ and $\pi_2$ that interact with the same partner $\pi^p$. Suppose $\pi_2$ explores the state space more broadly than $\pi_1$, which is reflected by: $H_{\rho^{\pi_1,\pi^p}}(S) \;<\; H_{\rho^{\pi_2,\pi^p}}(S)$. However, since $Z$ has fixed entropy $H(Z)$, maximizing the MI still yields
\begin{equation}
\max I(Z;S)_{\pi_1} \;=\; \max I(Z;S)_{\pi_2} \;=\; H(Z),
\end{equation}
meaning that $I(Z;S)$ alone does not distinguish policies with different exploration capacities. Thus, $I(Z;S)$ provides no extra information to prefer policies that better coordinate with the partner.

Furthermore, variational approximations of $H(Z \mid S)$ typically rely on neural networks (NN) trained via cross-entropy (CE) loss, which is well known to be biased toward spurious information in feature space \cite{Wei2023OnlinePL}. As a result, the agent can increase $I(Z;S)$ through local, repeatable perturbations that are largely agent-centric and do not improve coordination. Formally, maximizing $I(Z;S)$ does not prevent large conditional divergences, $\mathrm{KL}\big(\pi_{hi}(\cdot \mid s_1)\,\|\,\pi_{hi}(\cdot \mid s_2)\big), \quad s_1 \approx s_2,$ for states corresponding to similar partner behaviour, implying that high entropy does not necessarily ensure alignment with $\pi^p$. These observations motivate a more structured objective that conditions skill discovery on partner-relevant information and encourages skills to be meaningfully distinct. 
In particular, skill embeddings should capture partner behaviors, ensuring that discovered skills reflect collaborative interactions rather than agent-only perturbations.
\begin{figure}
\centering
\includegraphics[width=\linewidth]{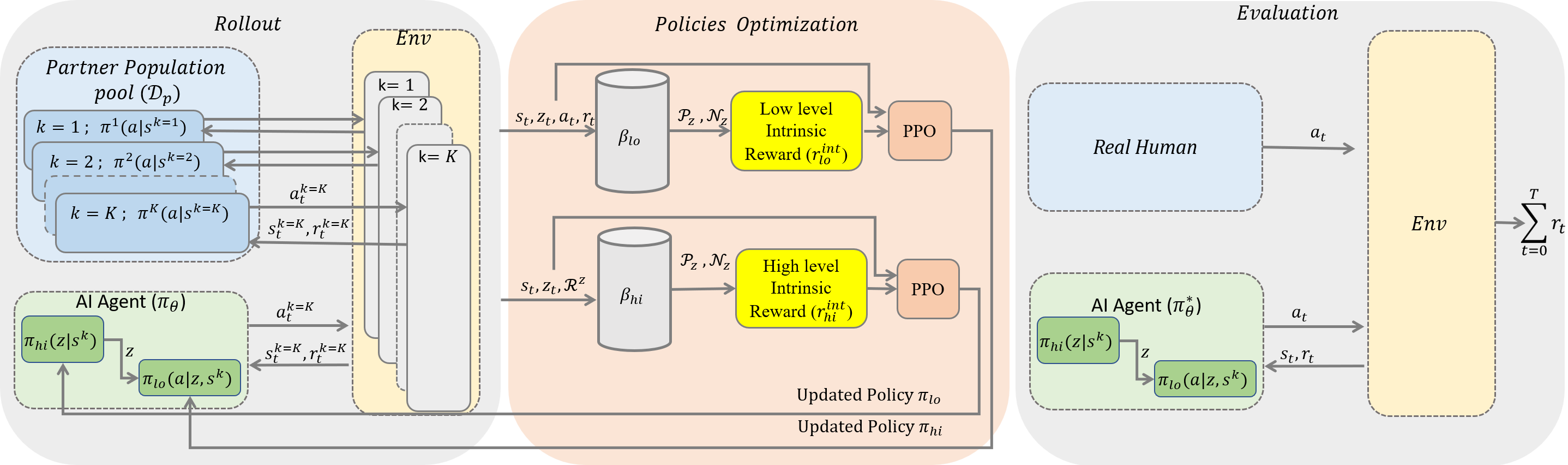}
\caption{
\textbf{Overview of PASD training and evaluation.}
\textbf{Left:} PASD is trained with $K$ parallel rollouts, each paired with a different partner sampled from the partner pool $\mathcal{D}_p$. After every episode, high-level and low-level trajectories are stored in buffers $\beta_h$ and $\beta_l$. These trajectories are used to form \emph{positive} pairs $\mathcal{P}_z$ (same skill across different partners) and \emph{negative} pairs $\mathcal{N}_z$ (different skills), enabling computation of the contrastive intrinsic rewards in Eq.~\ref{eq:intr_rew}. 
\textbf{Right:} The optimized policies $\pi_{hi}^{*}$ and $\pi_{lo}^{*}$ are evaluated with real human partners to measure collaborative performance.}
\label{fig:framework}
\end{figure}
\subsection{Partner-Adaptive Skills Discovery }
\textcolor{black}{In collaborative settings such as Overcooked \cite{carroll2019utility}, the behavior induced by a skill is shaped jointly by the agent and its partner. The same high-level skill may lead to different state transitions depending on whether the partner is fast, slow, or prioritizes different tasks. Thus, discovering meaningful skills requires capturing how a skill behaves across diverse partner policies, not just how the agent behaves in isolation. The objective in this section is to construct skill representations that remain consistent when interacting with behaviorally similar partners while remaining discriminative across different skills, ensuring that behaviorally distinct skills induce distinguishable interaction patterns. Figure~\ref{fig:framework} illustrates the overall PASD framework, showing both the training setup with partner interactions and the evaluation process with human partners.} To capture partner influence as a skill discriminability measure, consider collecting $K$ parallel rollouts under the joint dynamics of the skill-conditioned agent and the partner:
\begin{equation}
\tau^{(k)} \sim \rho^{\pi(\cdot\mid s,z),\pi^p \sim D^p}(\tau), \quad k=1,\dots,K,    
\end{equation}
where $\rho^{\pi,\pi^p}(\tau)$ denotes the trajectory distribution induced by the agent-partner interaction. Each rollout is segmented into $M$ sub-trajectories of length $L$:
\begin{equation}
\tau^{(k,j)} = \{s_{t_j}, a_{t_j}, \dots, s_{t_j+L}\}, \quad j=1,\dots,M,    
\end{equation}

and representative states are sampled from each sub-trajectory,  $s^{(k,j)} \sim \tau^{(k,j)}$.

\textbf{Assumption 2:}
We assume that distinct sub-trajectory views of the same skill encode a consistent partner-adaptive strategy, independent of which partner $\pi^p \sim \mathcal{D}_p$ is sampled, up to stochastic noise. 
In other words, for any two views $(k_1,j_1)\neq (k_2,j_2)$, the additional information about the skill identity provided by one view given the other is negligible:
\begin{equation}
I(S^{(k_1,j_1)}; Z \mid S^{(k_2,j_2)}) \approx 0, \quad \text{or equivalently} \quad S^{(k_1,j_1)} \perp\!\!\!\perp Z \mid S^{(k_2,j_2)}.
\end{equation}
Intuitively, once one view is observed, other views add little new information about the skill identity, reflecting reproducible partner-conditioned behavior across the partner population despite stochastic variations in trajectories. Under Assumption 2, the MI between sub-trajectory states can be used to discover useful partner-conditioned skills i.e., skills that are both distinct with diverse partner behavior and consistent across partners showing similar behavior:
\begin{equation} \label{eq:consistent_obj}
I(S^{(k_1,j_1)}; S^{(k_2,j_2)}) 
= I(S^{(k_1,j_1)}; S^{(k_2,j_2)}; Z) + I(S^{(k_1,j_1)}; S^{(k_2,j_2)} \mid Z, \pi^p),
\end{equation}
where the first term captures \emph{skill-discriminative information}, and the second term captures \emph{intra-skill consistency} across similar partner behaviors.

Direct computation of the MI objective in Equation (\ref{eq:consistent_obj}) is generally intractable.  Following prior work \cite{Oord2018RepresentationLW}, we approximate it using a tractable lower bound implemented via a contrastive objective over learned state embeddings $\phi(s)$. 
This approximation preserves the discriminability of skills across heterogeneous partner behaviors while maintaining consistency for similar partner behaviors, and can be directly interpreted as an intrinsic reward signal to shape skill representations. 

For each skill $z$, let the index set of its sub-trajectory views be  $\mathcal{P}_z \equiv \{(k,j) : \tau^{(k,j)}\}.$ Positive pairs are sampled from two distinct views $(k_1,j_1),(k_2,j_2)\in \mathcal{P}_z$, while negative samples are drawn from views of other skills,
$\mathcal{N}_z \equiv \bigcup_{z'\neq z} \mathcal{P}_{z'}$. We approximate MI using an InfoNCE-style contrastive loss \citep{guo2022online} over normalized embeddings $\phi(s)$, i.e., $\|\phi(s)\|_2 = 1$, so that similarities are measured on the unit hypersphere.  
For an anchor state $s$, with positive set $\mathcal{P}_z$ and negative set $\mathcal{N}_z$, per-anchor InfoNCE loss is: 
\begin{equation} \label{eq:infonce}
\mathcal{L}_{\mathrm{InfoNCE}}= - \frac{1}{|\mathcal{P}_z|} \sum_{s^+ \in \mathcal{P}_z} \log
\frac{\exp(\mathrm{sim}(\phi(s),\phi(s^+))/\tau)}
{\sum_{s' \in \mathcal{P}_z \cup \mathcal{N}_z} \exp(\mathrm{sim}(\phi(s),\phi(s'))/\tau)},
\end{equation}
where $\mathrm{sim}(\cdot,\cdot)$ is the cosine similarity and $\tau>0$ is a temperature parameter.

By the InfoNCE bound \cite{Oord2018RepresentationLW}, maximizing this reward increases a variational lower bound on the MI between the skill variable $Z$ and the state embeddings $\phi(S)$:
\begin{equation}
I(\phi(S); Z) \ge \log(N) - \mathcal{L}_{\mathrm{InfoNCE}},
\end{equation}
where $N = |\mathcal{N}_z|$ is the number of negatives.  The tightness of this approximation depends on the number of negative samples $N$ and the total number of skills $|\mathcal{Z}|$. Larger numbers of negatives and more diverse skills increase the quality of the lower bound, providing a stronger learning signal. This formulation ensures that the learned embeddings $\phi(s)$ capture both skill distinctiveness and consistency across partner behaviors. \textcolor{black}{The InfoNCE objective is applied over carefully constructed positive and negative pairs. Positive pairs consist of sub-trajectories generated by the same skill interacting with different partners, which encourages embeddings to be consistent across partner behaviors. Negative pairs come from sub-trajectories of other skills, ensuring embeddings are distinct for behaviorally different skills. By maximizing InfoNCE, the learned embeddings $\phi(s)$ capture patterns that are reproducible and conditioned on the partner, rather than arbitrary agent-centric state differences. This ensures that the discovered skills reflect meaningful partner-adaptive dynamics.}

\subsection{Contrastive Intrinsic Reward}

To facilitate the discovery of partner-conditioned skills, we derive an intrinsic reward by leveraging the InfoNCE objective. Specifically, the per-anchor InfoNCE probability, which measures similarity between states corresponding to the same skill relative to other skills, can be directly used as an intrinsic reward signal for both high-level and low-level policies. 

For each anchor state $s \in \mathcal{P}_z$, we compute a contrastive intrinsic reward as
\begin{equation}\label{eq:intr_rew}
r^{\mathrm{int}}(s) = \frac{1}{|\mathcal{P}_z|} \sum_{s^+ \in \mathcal{P}_z} 
\frac{\exp(\mathrm{sim}(\phi(s),\phi(s^+))/\tau)}
{\sum_{s' \in \mathcal{P}_z \cup \mathcal{N}_z} \exp(\mathrm{sim}(\phi(s),\phi(s'))/\tau)},
\end{equation}
where $\phi(s)$ denotes a normalized state embedding ($\|\phi(s)\|_2=1$), $\mathrm{sim}(\cdot,\cdot)$ is the cosine similarity, and $\tau>0$ is a temperature parameter. This intrinsic reward encourages the policy to select skills that are both discriminative across heterogeneous partner behaviors and consistent across sub-trajectory views corresponding to partners with similar behaviors. 

\begin{algorithm}[t]
\caption{PASD — Rollout and Intrinsic Reward Computation}
\label{alg:rollout}
\begin{algorithmic}[1]
\State \textbf{Input:} Partner population $\mathcal{D}_p$, high-level policy $\pi_{hi}(z|s)$, low-level policy $\pi_{lo}(a|s,z)$
\State \textbf{Parameters:} Intrinsic reward weight $\lambda$, rollout horizon $T$, skill segment length $L$
\State Initialize empty rollout buffers $\mathcal{B}_{hi}, \mathcal{B}_{lo}$

\For{each parallel rollout $k=1,\dots,K$}
    \State Sample partner policy $\pi^p \sim \mathcal{D}_p$
    \State Reset environment $s_0 \sim \rho_0$
    \State Initialize $t \gets 0$

    \While{$t < T$}
        \State Sample skill $z_t \sim \pi_{hi}(z|s_t)$
        \Repeat
            \State Sample low-level action $a_t \sim \pi_{lo}(a|s_t, z_t)$
            \State Sample partner action $a_t^p \sim \pi^p(a|s_t)$
            \State Step environment: $s_{t+1}, r_t \gets \text{EnvStep}(s_t, a_t, a_t^p)$
            \State Store $(s_t, z_t, a_t, r_t)$ in low-level buffer $\mathcal{B}_{lo}$
            \State Sample termination $b_t \sim \beta(z_t, s_{t+1})$
            \State $t \gets t+1$
        \Until{termination $b_t$ or $t \ge T$}
        \State Compute high-level segment reward $R^Z(s_{t_k}, z_t)$ as sum of extrinsic rewards
        \State Store $(s_{t_k}, z_t, R^Z)$ in high-level buffer $\mathcal{B}_{hi}$
    \EndWhile
\EndFor

\State Construct positive pairs $\mathcal{P}_z$ and negative pairs $\mathcal{N}_z$ from $\mathcal{B}_{hi}$ for each skill $z$
\State Compute contrastive intrinsic rewards $r^{int}(s)$ using InfoNCE (Eq.~\ref{eq:intr_rew})
\State Combine intrinsic and extrinsic rewards using Equations : (\ref{eq:low_int}, \ref{eq:final_hi} and \ref{eq:final_lo})
\State \textbf{Output:} High-level buffer $\mathcal{B}_{hi}$, low-level buffer $\mathcal{B}_{lo}$ with combined rewards
\end{algorithmic}
\end{algorithm}
\subsection{Overall Training Objective}

To effectively learn partner-adaptive skills, we integrate the intrinsic reward defined in Equation (\ref{eq:intr_rew}) with the extrinsic environment reward in both high-level and low-level objectives mentioned in Equations (\ref{eq:J_hi} and \ref{eq:J_lo}). For the high-level manager, the intrinsic reward is accumulated and normalized over each skill segment $[t_k, t_{k+1}-1]$:

\begin{equation} \label{eq:low_int}
\tilde{\mathcal{R}}^Z(s_{t_k}, z_k) = \frac{1}{t_{k+1}-t_k} \sum_{t=t_k}^{t_{k+1}-1} \Big( (1-\lambda) \, r(s_t, a_t, a^p_t) + \lambda \, r^{\mathrm{int}}(s_t) \Big),
\end{equation}

where $\lambda \in [0,1]$ controls the relative weighting of intrinsic and extrinsic rewards.  
The corresponding high-level objective is

\begin{equation}\label{eq:final_hi}
\tilde{J}_{hi} = \mathbb{E}\Bigg[\sum_{k=0}^{K-1} \gamma^{t_k} \tilde{\mathcal{R}}^Z(s_{t_k}, z_k) \Bigg].
\end{equation}
 
In the early phase of training, the intrinsic reward dominates, promoting exploration of diverse skill patterns and capturing variations in partner behaviors across different rollouts. As training progresses, the influence of the extrinsic reward gradually increases, guiding the high-level manager to refine skill selection toward maximizing task returns while maintaining consistency and discriminability across partner-conditioned interactions. For the low-level controller, the intrinsic reward is applied at each timestep:

\begin{equation} \label{eq:final_lo}
\tilde{J}_{lo} = \mathbb{E}\Bigg[\sum_{t=t_k}^{t_{k+1}-1} \gamma^{t-t_k} \Big( r(s_t, a_t, a^p_t) + \lambda \, r^{\mathrm{int}}(s_t) \Big)\Bigg].
\end{equation}

Initially, the intrinsic reward drives the low-level policy to produce diverse and disentangled action distributions for each skill, capturing the variability in partner behaviors across different rollouts. As training progresses, the extrinsic reward gradually increases in influence, aligning these action distributions with task objectives while preserving the discriminability and consistency of behaviors for partners with similar tendencies. 
Both high-level and low-level objectives are optimized with the PPO algorithm, using the rewards defined in Equations~\ref{eq:final_hi} and~\ref{eq:final_lo}. Detailed pseudocode describing the rollout procedure and the policy optimization steps of PASD is provided in Algorithm~\ref{alg:rollout} and Algorithm~\ref{alg:ppo}, respectively. 

\begin{algorithm}[t]
\caption{PASD: Hierarchical Policy Update (High-level, Low-level, Termination)}
 \label{alg:ppo}
\begin{algorithmic}[1]
\State \textbf{Input:} High-level buffer $\mathcal{B}_{hi}$, low-level buffer $\mathcal{B}_{lo}$, high-level policy $\pi_{hi}$, low-level policy $\pi_{lo}$, termination policy $\beta$
\State \textbf{Parameters:} PPO clipping $\epsilon$, discount $\gamma$, GAE $\lambda_{GAE}$, learning rate $\alpha$

\For{each gradient update iteration}
    \State Compute high-level advantages $\hat{A}^h_t$ from $\mathcal{B}_{hi}$ using GAE
    \State Compute low-level advantages $\hat{A}^l_t$ from $\mathcal{B}_{lo}$ using GAE
    \State Compute termination advantages $\hat{A}^\beta_t$ from $\mathcal{B}_{hi}$ or $\mathcal{B}_{lo}$

    \State Compute PPO ratio $r^h_t(\theta) = \frac{\pi_{hi,\theta}(z_t|s_t)}{\pi_{hi,\theta_\text{old}}(z_t|s_t)}$
    \State Compute clipped PPO loss with entropy: $L^h = - \mathbb{E}\Big[\min(r^h_t \hat{A}^h_t, \text{clip}(r^h_t, 1-\epsilon, 1+\epsilon) \hat{A}^h_t)  \Big]$

    \State Compute PPO ratio $r^l_t(\theta) = \frac{\pi_{lo,\theta}(a_t|s_t,z_t)}{\pi_{lo,\theta_\text{old}}(a_t|s_t,z_t)}$
    \State Compute clipped PPO loss with entropy:
    $L^l = - \mathbb{E}\Big[\min(r^l_t \hat{A}^l_t, \text{clip}(r^l_t, 1-\epsilon, 1+\epsilon) \hat{A}^l_t)  \Big]$

    \State Compute PPO ratio $r^\beta_t(\theta) = \frac{\beta_\theta(b_t|s_t,z_t)}{\beta_{\theta_\text{old}}(b_t|s_t,z_t)}$
    \State Compute clipped PPO loss with entropy: $L^\beta = - \mathbb{E}\Big[\min(r^\beta_t \hat{A}^\beta_t, \text{clip}(r^\beta_t, 1-\epsilon, 1+\epsilon) \hat{A}^\beta_t)  \Big]$

    \State Update parameters $\theta$ of $\pi_{hi}, \pi_{lo}, \beta$ using $L^h + L^l + L^\beta$
\EndFor

\State \textbf{Return:} Updated policies $\pi_{hi}, \pi_{lo}, \beta$
\end{algorithmic}
\end{algorithm}

\section{Experiments}

\begin{figure*}
\centering
\includegraphics[width = \textwidth]{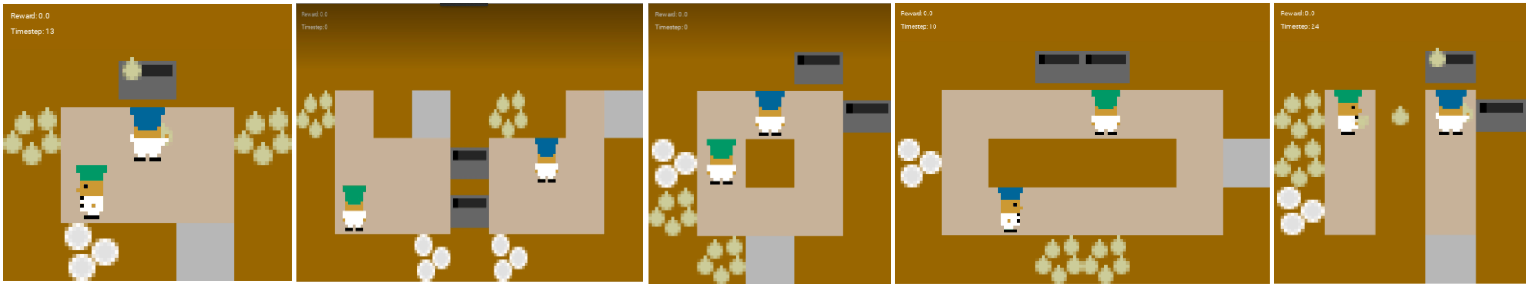}
\caption{The five standard Overcooked layouts (left to right): Cramped Room, Asymmetric Advantages, Coordination Ring, Counter Circuit and Forced Coordination.}
\label{fig:layouts}
\end{figure*}
\subsection{Experimental Details}
\paragraph{Environment:}
Following existing works \citep{strouse2021collaborating, loo2023hierarchical, yu2023learning, yang2023hierarchical}, we adopt the Overcooked-AI \cite{carroll2019utility} as our testbed. Overcooked-AI is a two-player cooperative benchmark derived from the Overcooked game \cite{ghosttown2016overcooked}, in which agents collaboratively complete a soup preparation task. Agents must pick onions, place them in the pot, wait for the soup to cook, and then deliver the completed soup to the serving station, with each successful delivery yielding a reward of +20. The goal is to maximize team reward within a fixed episode horizon. We evaluate across five standard layouts, \emph{Cramped Room}, \emph{Asymmetric Advantages}, \emph{Coordination Ring}, \emph{Forced Coordination}, and \emph{Counter Circuit}, illustrated in Figure~\ref{fig:layouts}. These layouts present diverse coordination challenges and collectively provide a widely adopted benchmark for studying partner-adaptive behaviors. \textcolor{black}{Overcooked-AI is fully observable, making it a suitable testbed where coordination challenges arise solely from partner behavior rather than partial observability \cite{ yu2023learning, yang2023hierarchical}}. For a detailed discussion of layout-specific demands, see Appendix~\ref{appendix:layouts}.

\paragraph{Diverse Self-Play Partner Population:} \label{sec:diverse_partner}
 For effective coordination with novel partners and humans, the AI agent is trained with a diverse partner population, where each partner has a unique play style and skill level. Following prior work~\cite{strouse2021collaborating, pmlr-v139-lupu21a, loo2023hierarchical}, we construct a heterogeneous policy pool. The pool consists of 16 agents trained via self-play with PPO, varying in play style and skill level. Diverse play styles are encouraged using a negative Jensen–Shannon Divergence~\citep{loo2023hierarchical}, and varying skill levels are included via intermediate checkpoints~\citep{strouse2021collaborating}. During training, a partner is uniformly sampled from the heterogeneous population each episode. For evaluation with novel AI partners, a separate disjoint population of the same size is trained using the same procedure.
 
\paragraph{Baselines }
We compare PASD against standard baselines including FCP \citep{strouse2021collaborating}, DIAYN \citep{eysenbach2019diversity}, and HiPT \citep{loo2023hierarchical}. Each method is trained and evaluated with the identical set of diverse partner populations introduced earlier. FCP trains the adaptive policy directly with PPO, whereas DIAYN, HiPT, and PASD adopt HRL where high-level and low-level policies are jointly optimized via the option-critic framework \citep{SUTTON1999181}. DIAYN uses a two-stage process where skills are first acquired through intrinsic rewards and then fine-tuned for task performance, while HiPT and PASD train both levels of the hierarchy in parallel.
\paragraph{Implementation Details:}
We train all methods for $10^7$ steps using 30 parallel rollouts with a horizon length of 400. For PASD, the weighting coefficient $\lambda$ is linearly annealed from 1.0 to 0.05 during training.
 
Both low-level and high-level policies share a backbone network that consists of three convolution layers, two fully connected layers, and a recurrent LSTM layer. The network is then split into separate heads for low-level action and value prediction, and for high-level skill and value estimation. The discrete skill variable $z$ is set to dimension 6 for all layouts except \textit{Forced Coordination}, where it is set to 5. Additional hyperparameter details are provided in Appendix~\ref{appendix:implementation}.
\begin{table}[t]
\centering
\caption{Total mean reward (Mean ± Std) across three versions of each evaluation partner (early, intermediate, final checkpoint) and both starting positions.}
\label{tab:sp_pop}
\begin{tabular}{lccccc}
\toprule
Method & Cramped Room & Asym. Adv. & Coord. Ring & Counter Circuit & Forced Coord. \\
\midrule
FCP   & 137.7 ± 1.0 & 90.6 ± 1.0 & 83.9 ± 5.9 & 51.3 ± 5.0 & 36.7 ± 14.4 \\
DIAYN & 33.8 ± 6.4  & 1.5 ± 0.7  & 22.5 ± 6.3 & 1.2 ± 1.0  & 1.3 ± 0.0   \\
HiPT  & 117.9 ± 4.4 & 86.2 ± 0.9 & 96.0 ± 1.3 & 38.1 ± 5.3 & 35.6 ± 13.0 \\
PASD  & \textbf{165.8 ± 10.0} & \textbf{145.8 ± 9.6} & \textbf{101.3 ± 8.5} & \textbf{57.37 ± 2.9} & \textbf{46.87 ± 12.3} \\
\bottomrule
\end{tabular}
\end{table}
\subsection{Results}
We organize our results into three evaluation categories based on partner type. First, the agent is paired with a diverse self-play population (Section~\ref{sec:diverse_partner}) to test adaptation to novel AI behaviors. Second, we evaluate with human proxy models trained via behavior cloning on human–human trajectories, providing a closer approximation of human collaboration. Third, we validate performance through a controlled human-subject study to assess real human-AI coordination.

\begin{table}[]
\centering
\caption{Total mean reward across different layouts when paired with a Behaviour Cloning (BC) partner.}
\label{tab:bc_partner}
\begin{tabular}{lccccc}
\toprule
Method & Cramped Room & Asym. Adv. & Coord. Ring & Counter Circuit & Forced Coord. \\
\midrule
FCP     & 118.75 & 80.00  & 79.38  & 38.13  & 30.75 \\
DIAYN   & 40.00  & 0.00   & 26.25  & 1.25   & 6.30 \\
HIPT    & 93.13  & 66.25  & 77.50  & 35.00  & 25.20 \\
PASD & \textbf{150.00 }& \textbf{112.50} & \textbf{105.63} & \textbf{44.38}  & \textbf{43.8} \\
\bottomrule
\end{tabular}
\end{table}
\begin{table}[]
\centering
\caption{\textcolor{black}{Total mean reward (Mean ± Standard Deviation) achieved by human participants when paired with HiPT and PASD across different Overcooked layouts.}}
\label{tab:hm_partner}
\begin{tabular}{lccccc}
\toprule
Method & Cramped Room & Asym. Adv. & Coord. Ring & Counter Circuit & Forced Coord. \\
\midrule
HiPT  & $80.00 \pm 16.34$  & $136.36 \pm 20.36$ & $46.0 \pm 12.36$ & $40.00 \pm 15.81$  & $20 \pm 0.0$ \\
PASD  & \textbf{$118.18 \pm 12.68$} & \textbf{$198.18 \pm 19.41$} & \textbf{$60.0 \pm 08.92$} & \textbf{$62.5 \pm 10.95$} & \textbf{$35.0 \pm 10.00$} \\
\bottomrule
\end{tabular}
\end{table}
\begin{figure*}[]
\centering
\begin{subfigure}{0.32\textwidth}
    \includegraphics[width=\linewidth]{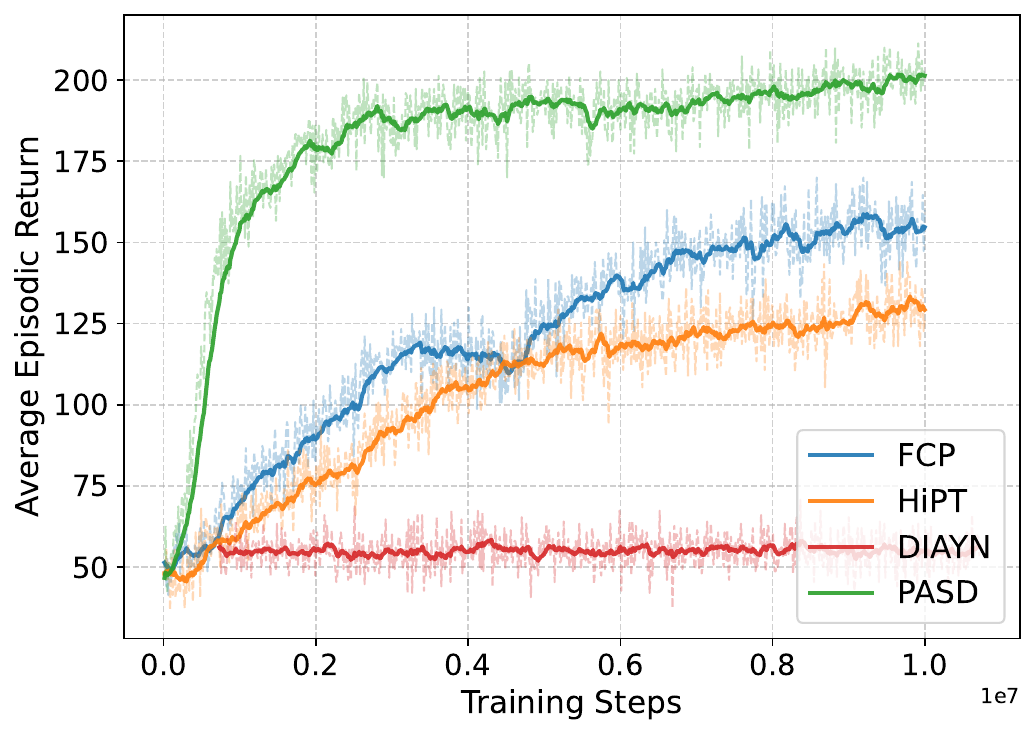}
    \caption{Cramped Room}
    \label{fig:cramped_room}
\end{subfigure}
\begin{subfigure}{0.32\textwidth}
    \includegraphics[width=\linewidth]{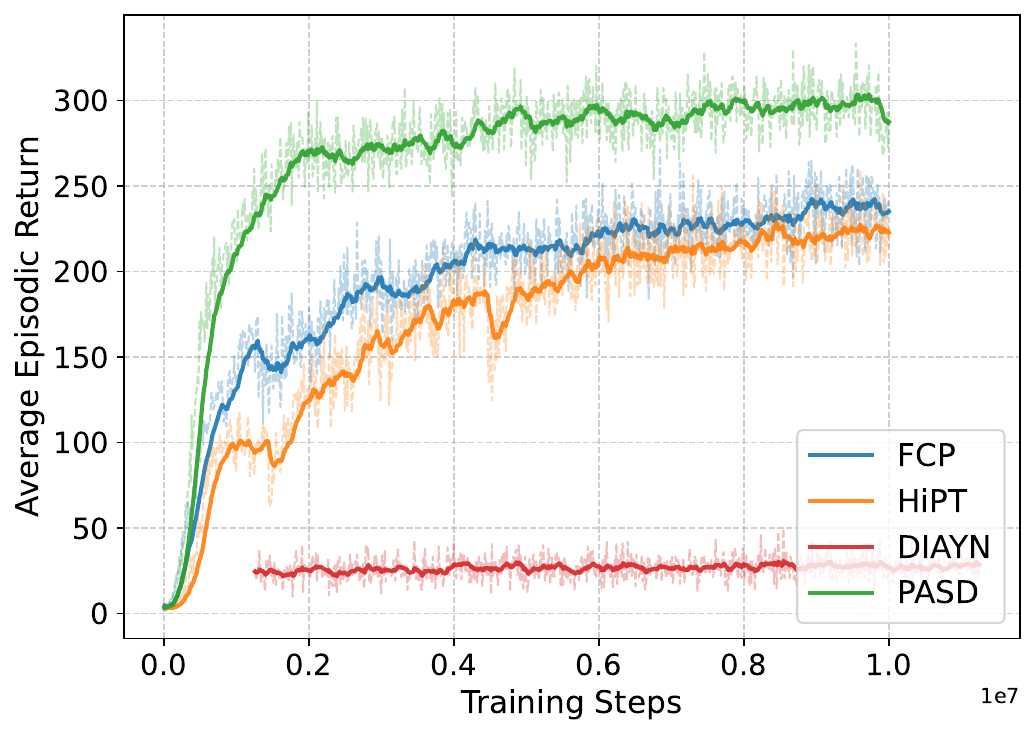}
    \caption{Asymmetric Advantages}
    \label{fig:asymmetric_advantages}
\end{subfigure}
\begin{subfigure}{0.32\textwidth}
    \includegraphics[width=\linewidth]{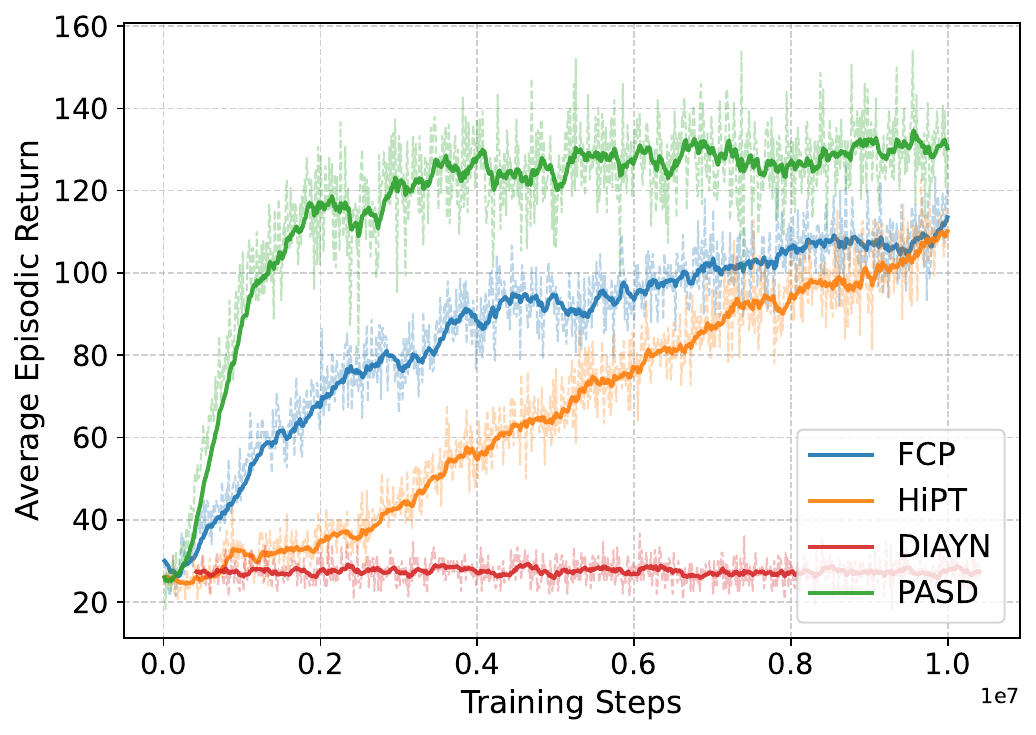}
    \caption{Coordination Ring}
    \label{fig:coordination_ring}
\end{subfigure}

\begin{subfigure}{0.32\textwidth}
    \includegraphics[width=\linewidth]{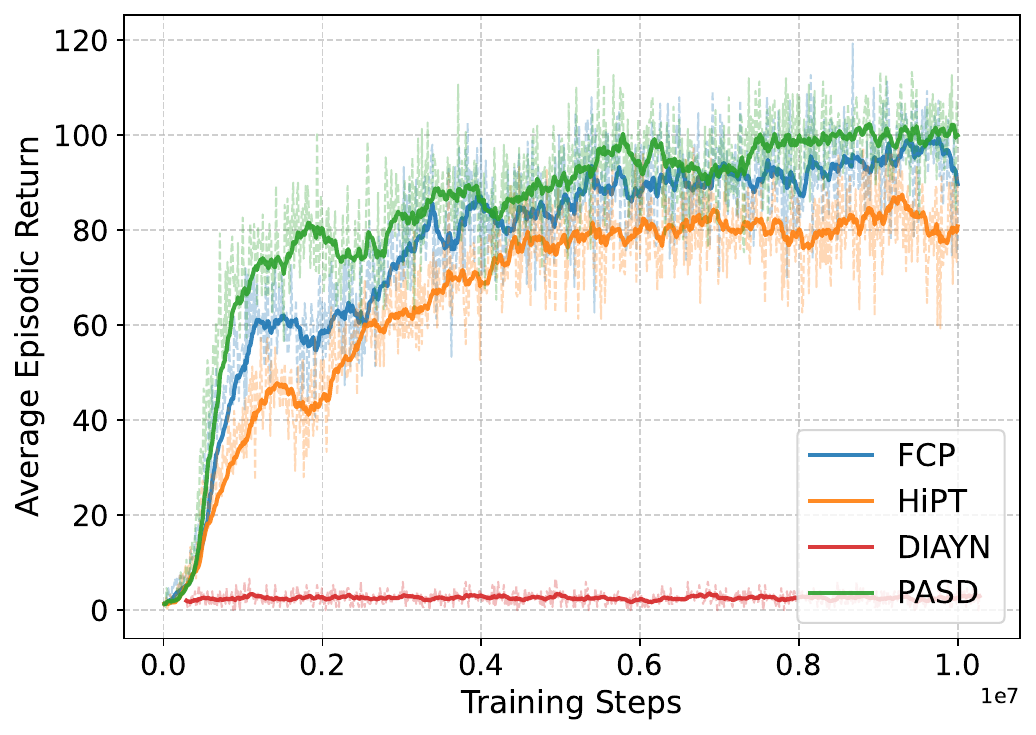}
    \caption{Counter Circuit}
    \label{fig:counter_circuit}
\end{subfigure}
\begin{subfigure}{0.32\textwidth}
    \includegraphics[width=\linewidth]{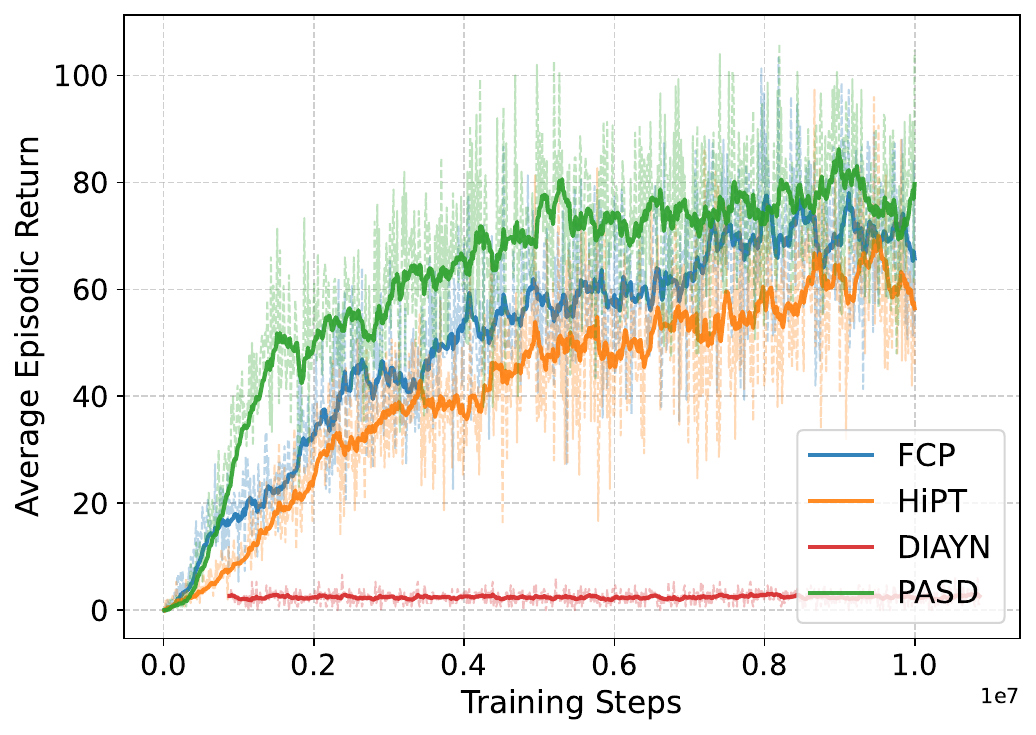}
    \caption{Forced Coordination}
    \label{fig:forced_coordination}
\end{subfigure}

\caption{Average episodic return during training across 30 parallel rollout environments.}
\label{fig:hipt_vs_riayn_all}
\end{figure*}

\begin{figure}
\centering
\begin{subfigure}{0.24\textwidth}
    \includegraphics[width=\linewidth]{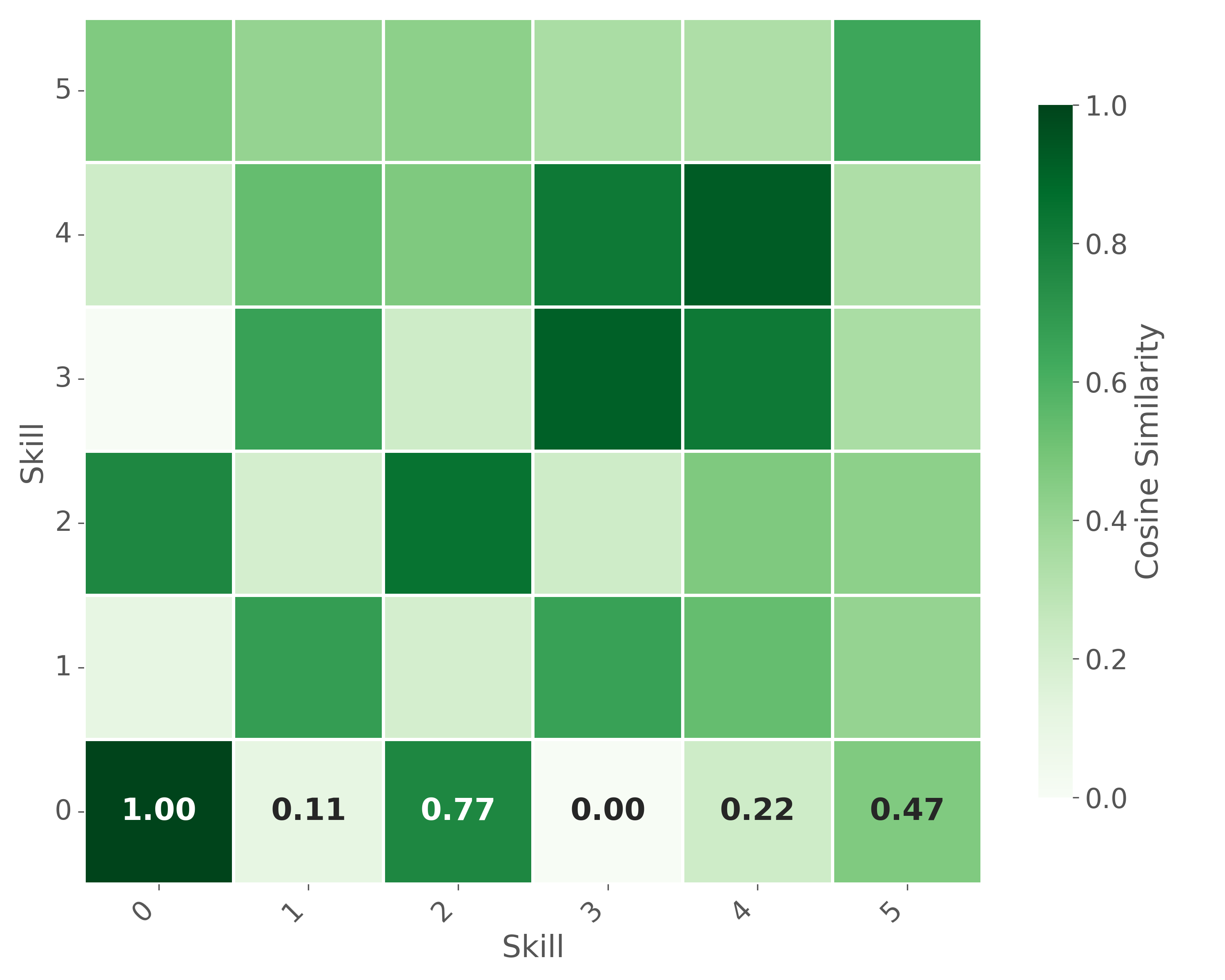}
    \caption{Cramped Room: Human}
    \label{fig:cramped_human_cossim}
\end{subfigure}\hfill
\begin{subfigure}{0.24\textwidth}
    \includegraphics[width=\linewidth]{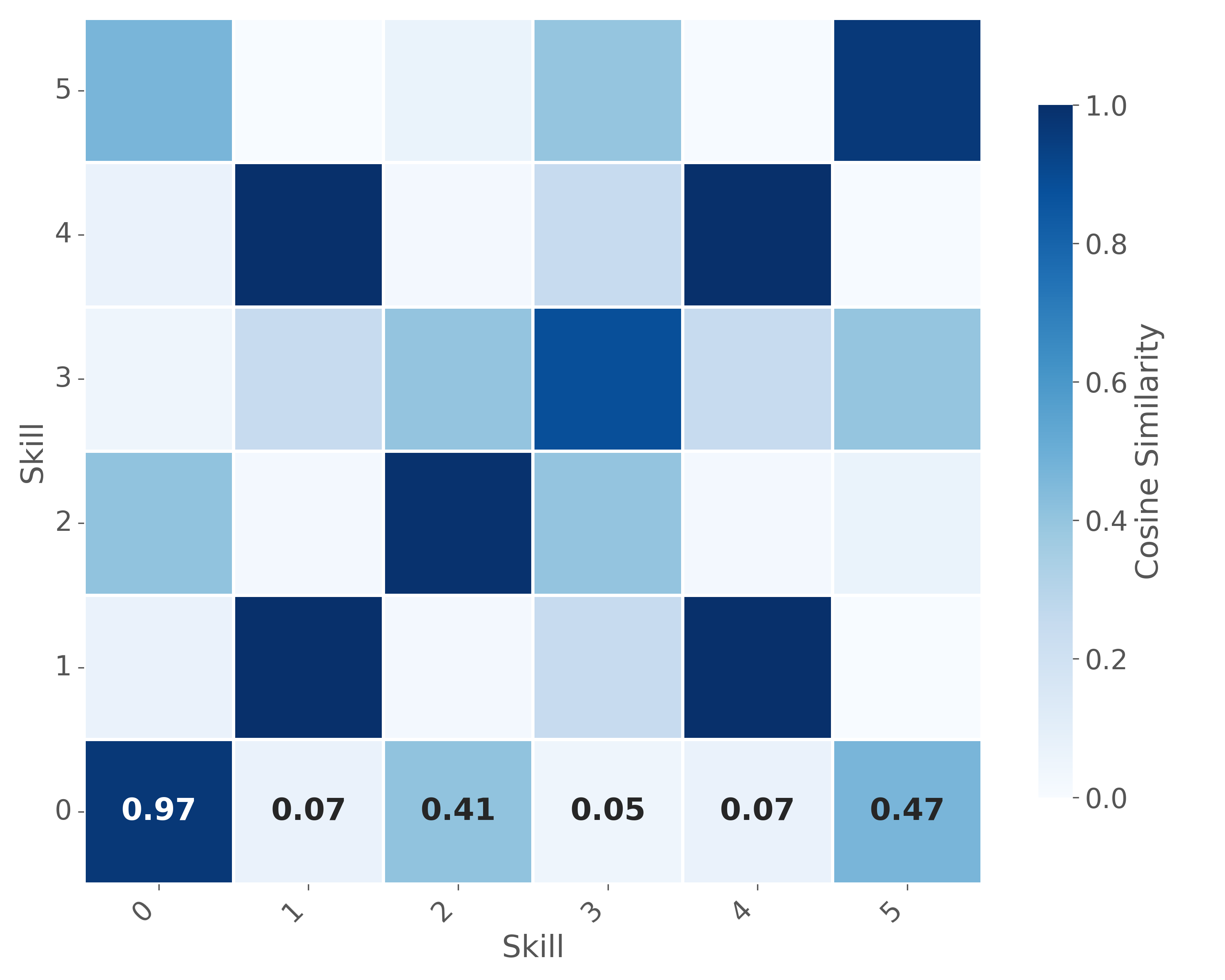}
    \caption{Cramped Room: AI-Agent}
    \label{fig:cramped_ai_cossim}
\end{subfigure}\hfill
\begin{subfigure}{0.24\textwidth}
    \includegraphics[width=\linewidth]{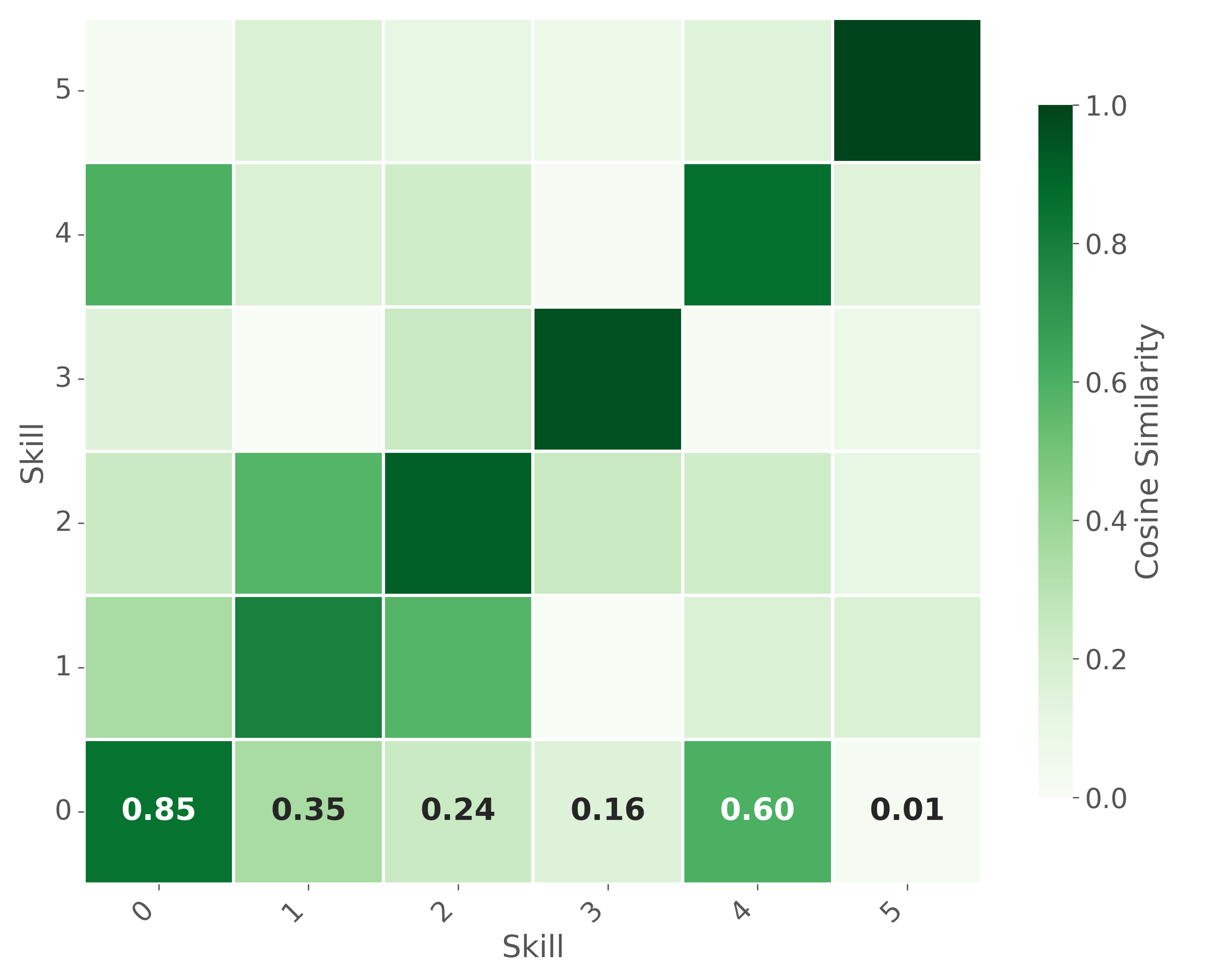}
    \caption{Coordination Ring: Human}
    \label{fig:coordination_human_cossim}
\end{subfigure}\hfill
\begin{subfigure}{0.24\textwidth}
    \includegraphics[width=\linewidth]{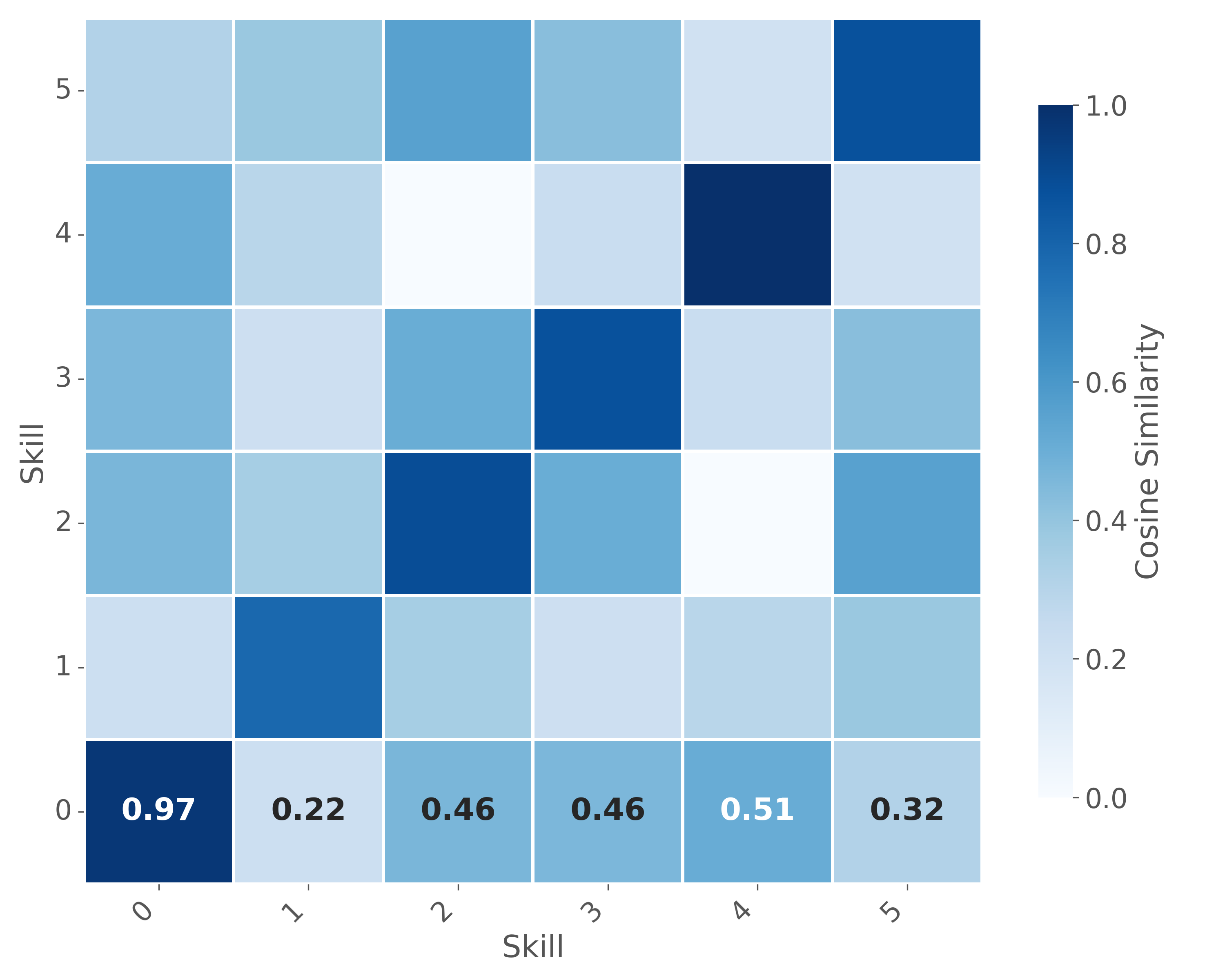}
    \caption{Coordination Ring: AI-Agent}
    \label{fig:coordination_ai_cossim}
\end{subfigure}
\caption{\textcolor{black}{Cosine-similarity heatmaps computed over sequence-level embeddings for the Cramped Room and Coordination Ring layouts. High values along the diagonal indicate strong intra-skill consistency, while low off-diagonal values reflect clear inter-skill separation}}
\label{fig:pasd_skill_anal0}
\end{figure}
\begin{figure}
\centering

\begin{subfigure}{0.45\textwidth}
    \includegraphics[width=\linewidth]{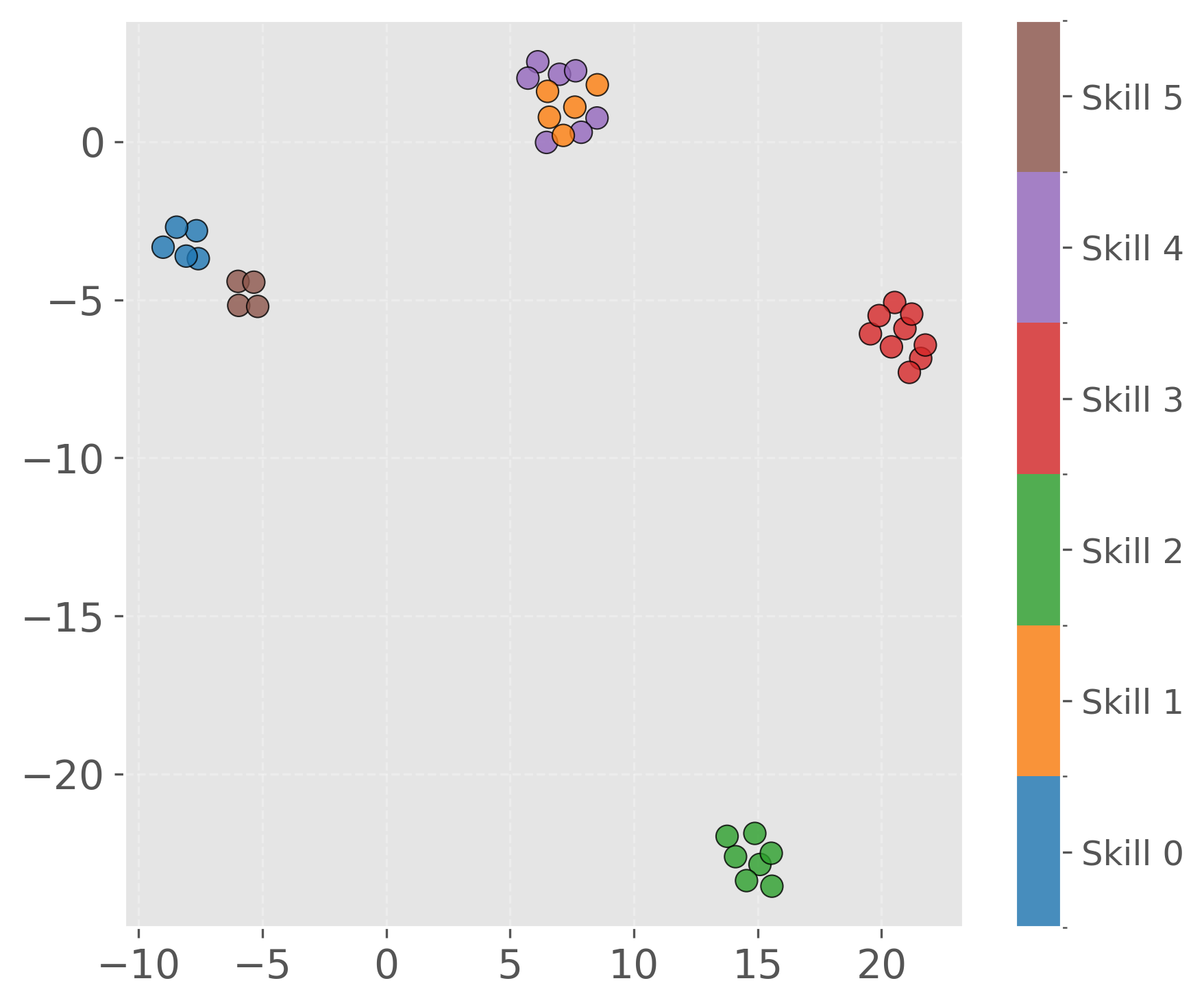}
    \caption{Cramped Room}
    \label{fig:cramped_ai_umap}
\end{subfigure}\hfill
\begin{subfigure}{0.45\textwidth}
    \includegraphics[width=\linewidth]{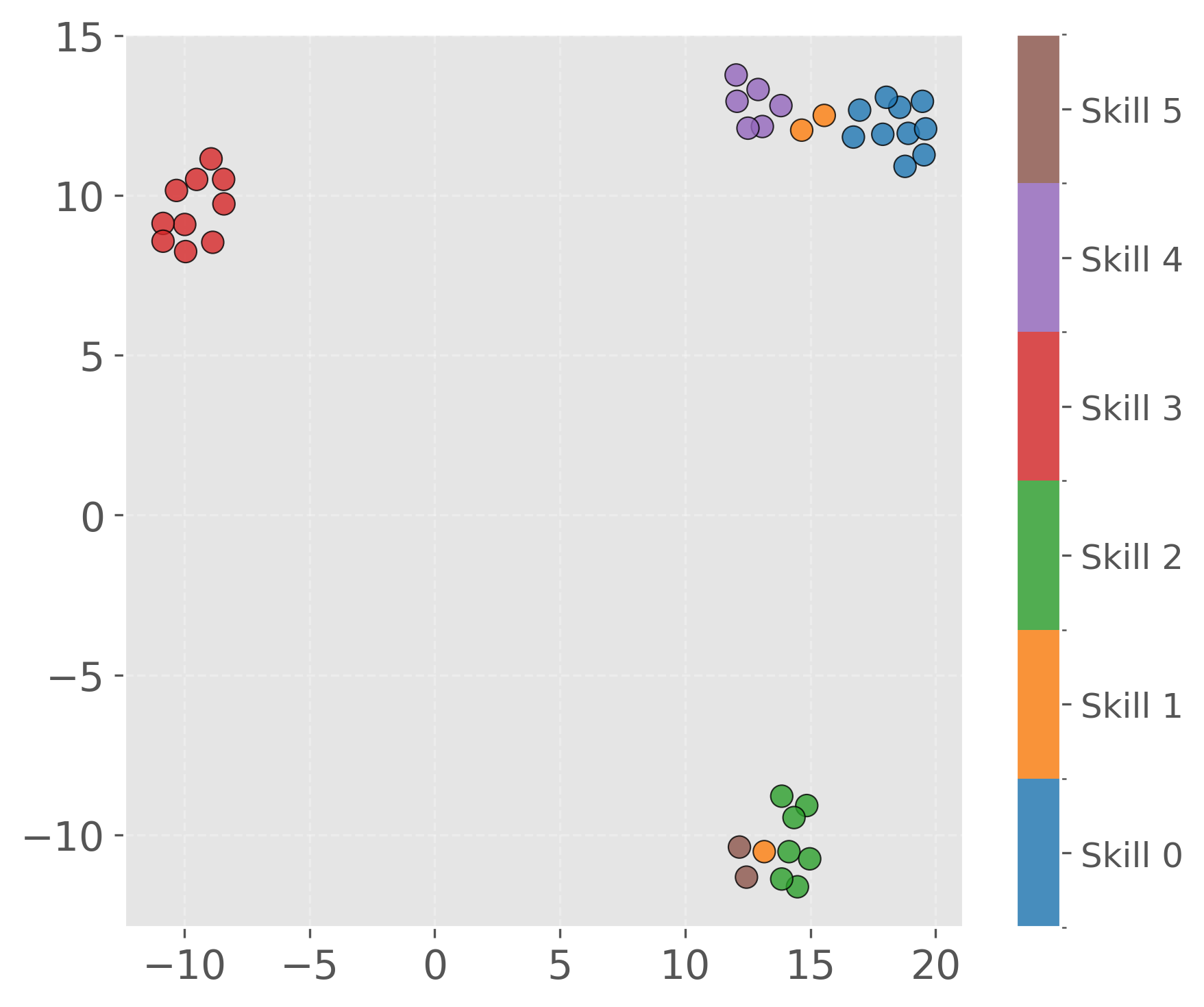}
    \caption{Coordination Ring}
    \label{fig:coordination_ai_umap}
\end{subfigure}
\caption{\textcolor{black}{Two-dimensional UMAP projections of sequence-level embeddings for Cramped Room and Coordination Ring layouts. Each point represents a skill segment, colored according to the active skill. Compact clusters for individual skills illustrate intra-skill consistency, while separation between clusters indicates inter-skill disentanglement.}}
\label{fig:pasd_skill_umap}
\end{figure}

\begin{figure}
\centering
\begin{subfigure}{0.32\textwidth}
    \includegraphics[width=\linewidth]{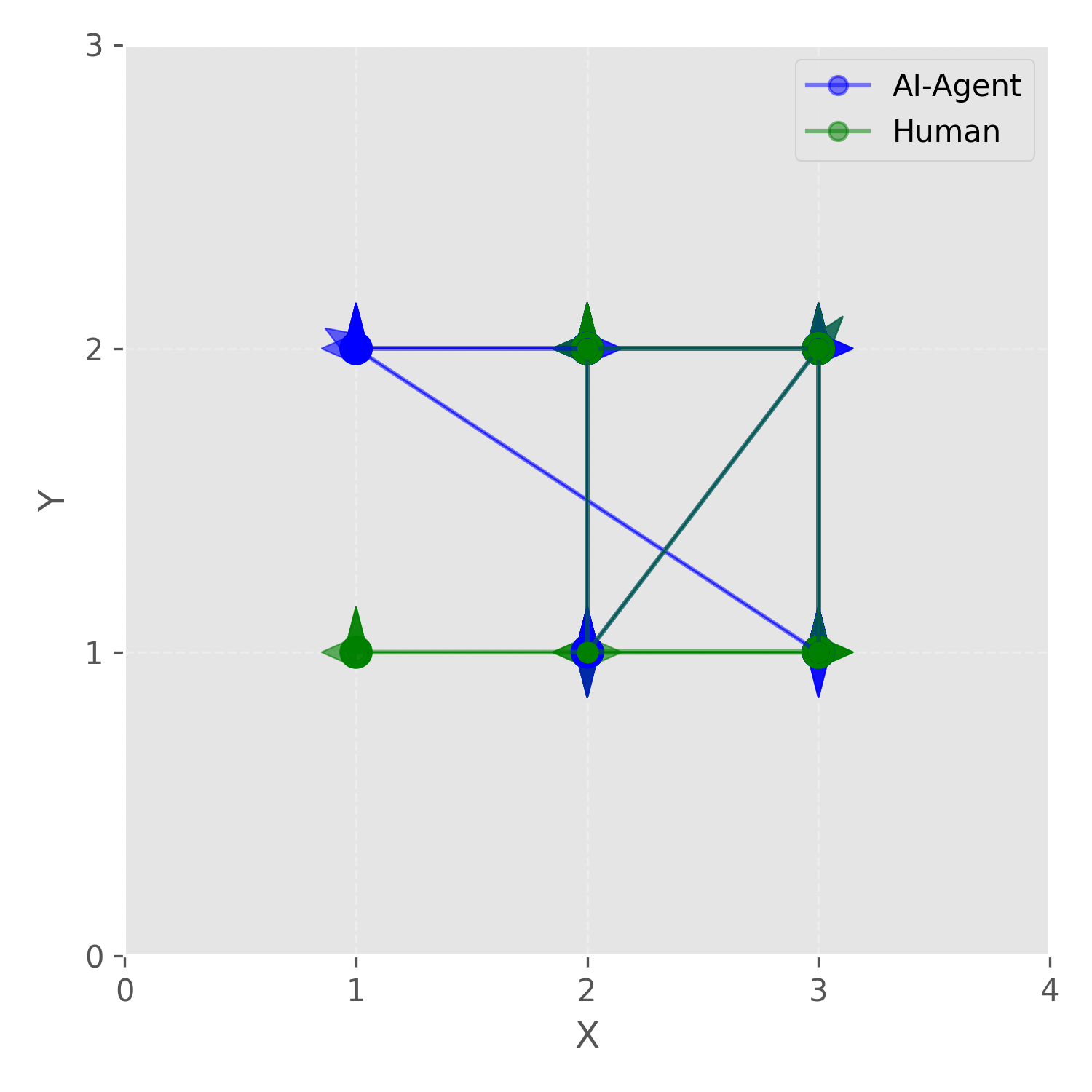}
    \caption{Skill:0}
    \label{fig:cramped_skill_0}
\end{subfigure}\hfill
\begin{subfigure}{0.32\textwidth}
    \includegraphics[width=\linewidth]{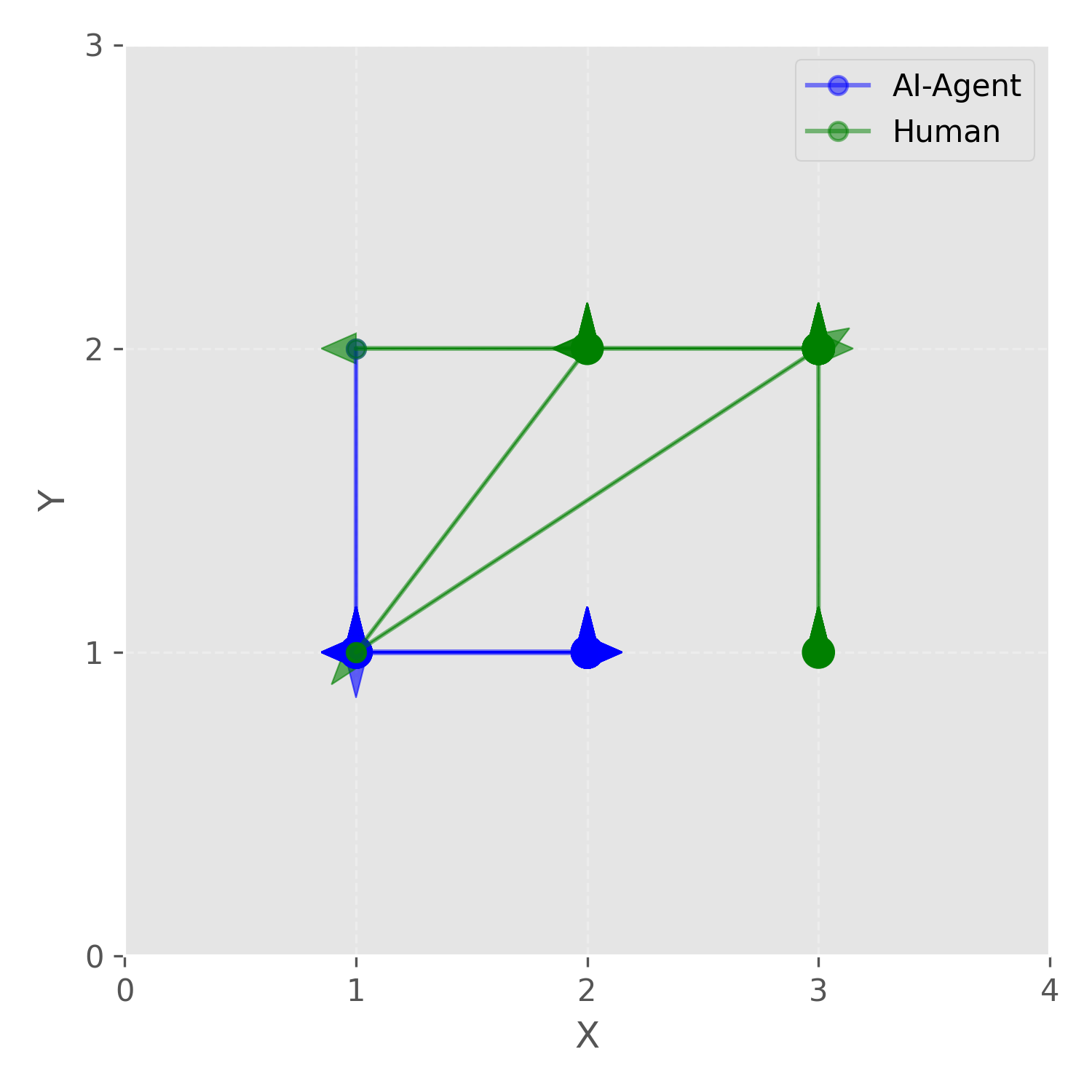}
    \caption{Skill:1}
    \label{fig:cramped_skill_1}
\end{subfigure}\hfill
\begin{subfigure}{0.32\textwidth}
    \includegraphics[width=\linewidth]{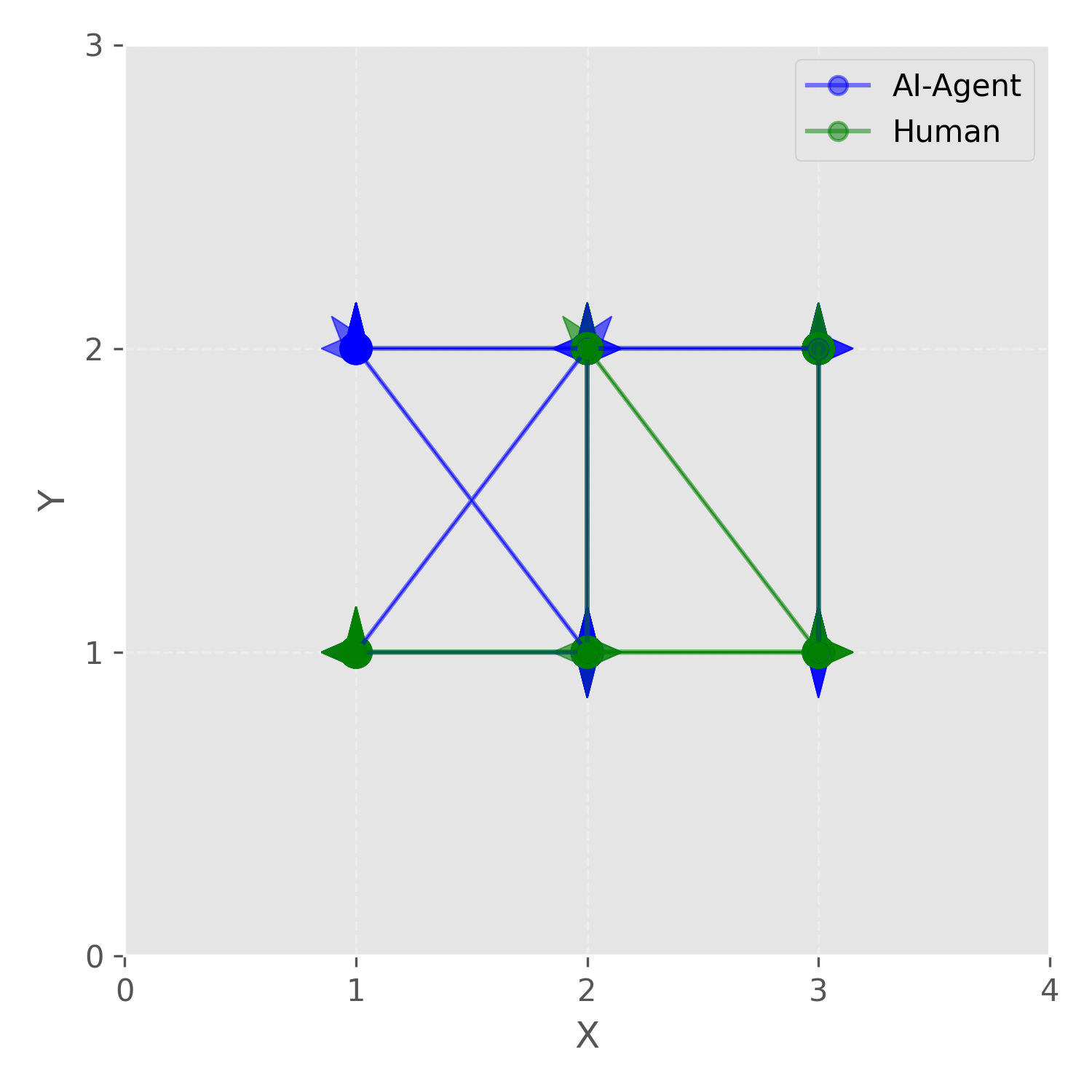}
    \caption{Skill:2}
    \label{fig:cramped_skill_2}
\end{subfigure}\hfill
\begin{subfigure}{0.32\textwidth}
    \includegraphics[width=\linewidth]{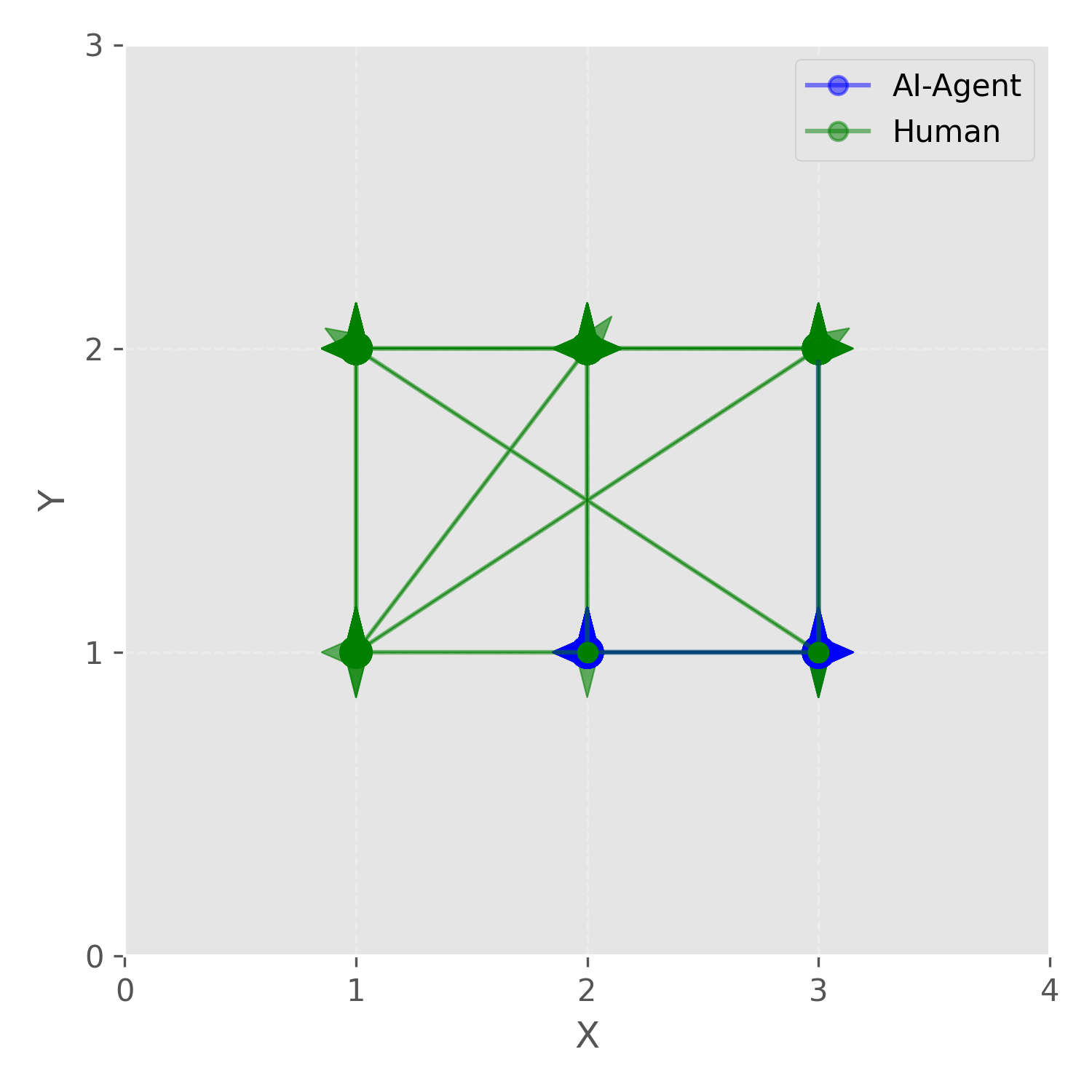}
    \caption{Skill:3}
    \label{fig:cramped_skill_3}
\end{subfigure}\hfill
\begin{subfigure}{0.32\textwidth}
    \includegraphics[width=\linewidth]{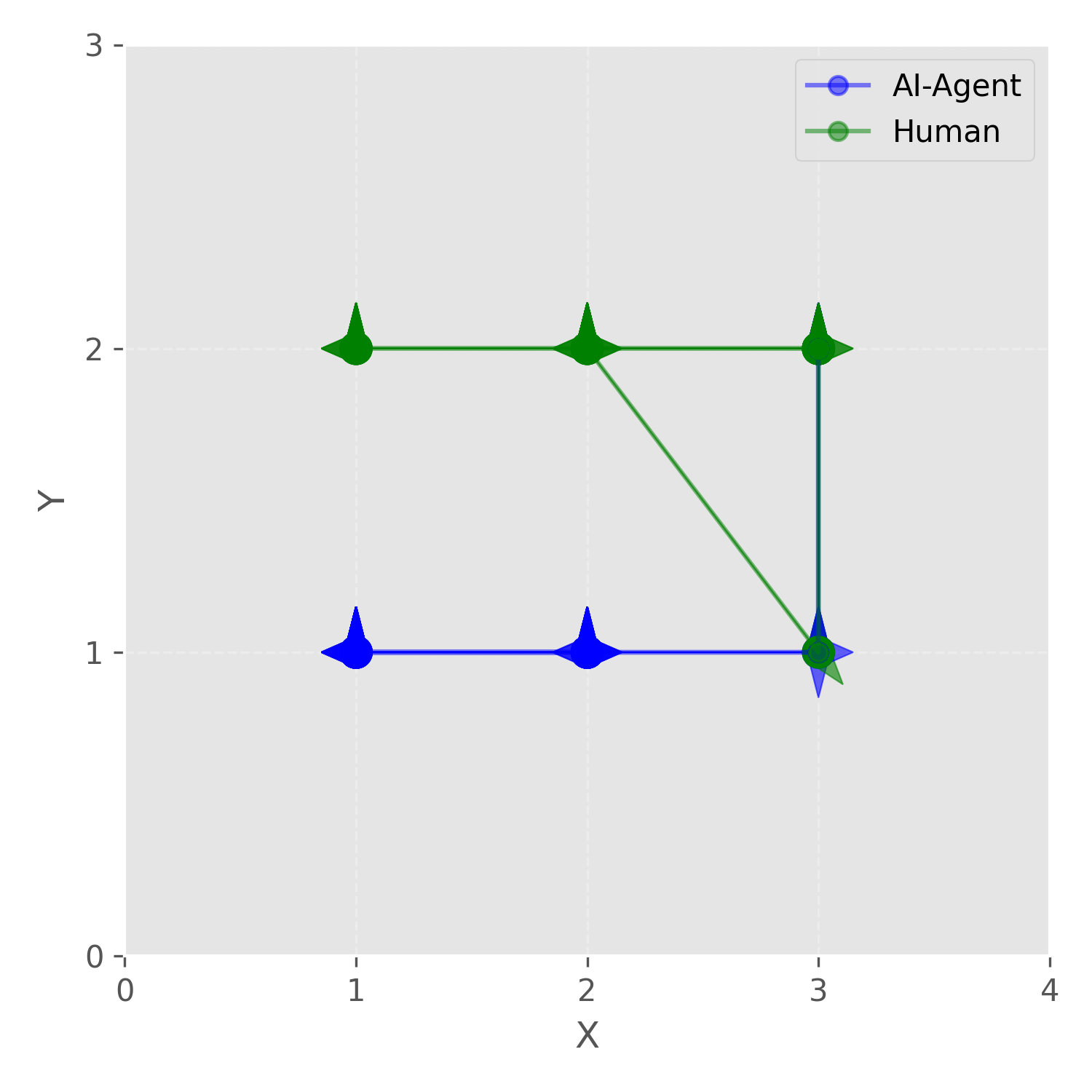}
    \caption{Skill:4}
    \label{fig:cramped_skill_4}
\end{subfigure}\hfill
\begin{subfigure}{0.32\textwidth}
    \includegraphics[width=\linewidth]{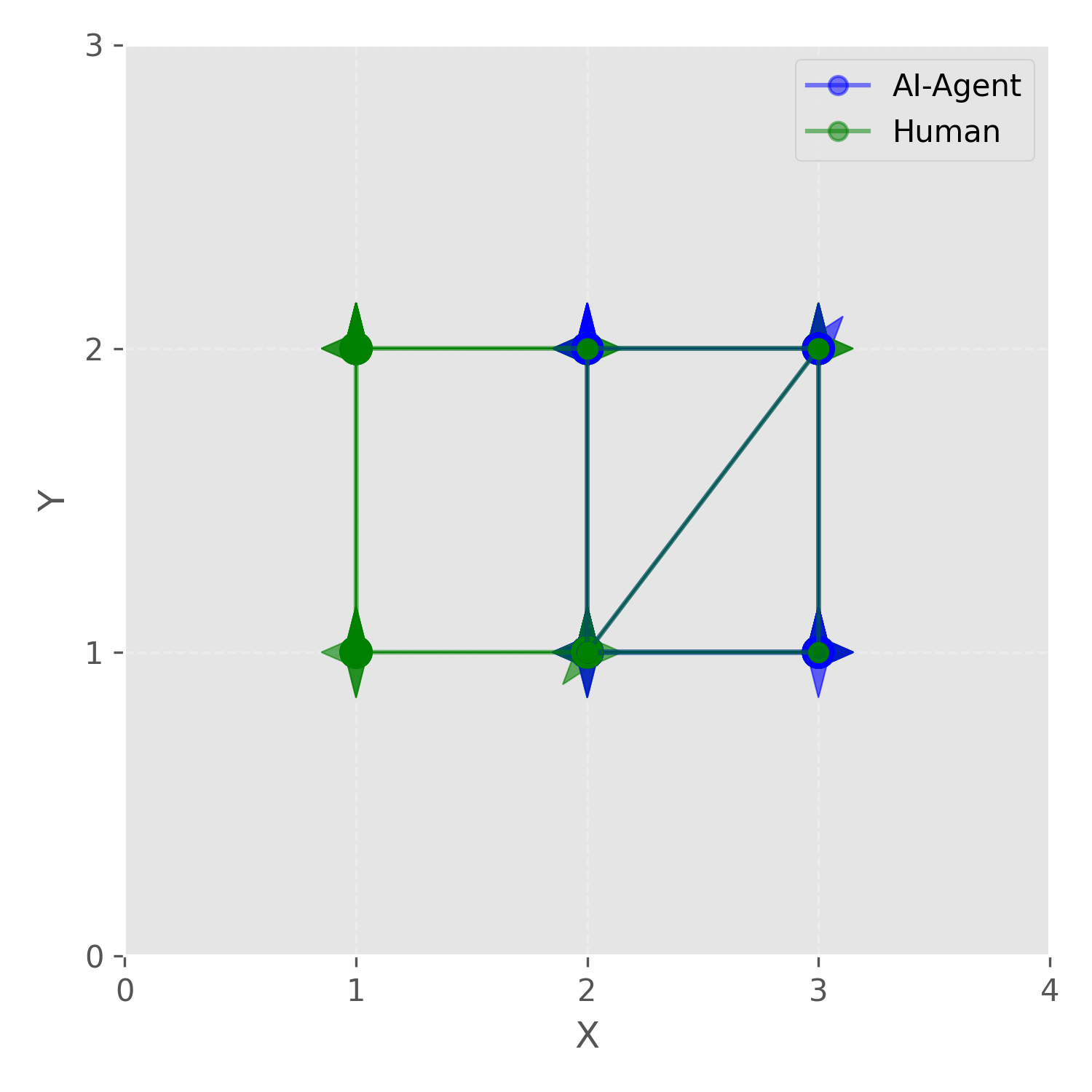}
    \caption{Skill:5}
    \label{fig:cramped_skill_5}
\end{subfigure}
\caption{\textcolor{black}{Human and AI-agent trajectories in the Cramped Room layout for different skill activations. Arrows indicate movement directions and circles denote interaction or stay actions. Each skill corresponds to a distinct human sub-trajectory and a skill-conditioned AI response.}}
\label{fig:pasd_skill_traj}
\end{figure}
\paragraph{Evaluation with Self-Play Partner Population:}
\label{sec:eval_novel}
We first evaluate all methods using the heterogeneous partner population, organized into three sets,early-stage, intermediate, and fully trained policies, covering a spectrum of partner proficiency from beginner to advanced. This population serves as a practical proxy for varied human collaborative behaviors \citep{strouse2021collaborating, yu2023learning}. For each set, we report the mean episodic return across all partners and starting positions, with overall performance summarized as mean ± standard deviation across the three sets (Table~\ref{tab:sp_pop}). Performance varies with layout difficulty. DIAYN performs poorly due to spurious variations disrupting skill learning. HiPT and FCP perform better but remain sensitive to redundant state information. In contrast, PASD robustly captures partner-relevant behaviors, avoids spurious variations, and achieves the highest returns across all layouts. Figures~\ref{fig:cramped_room}--\ref{fig:forced_coordination} show average returns over 30 rollouts, illustrating that PASD converges faster and maintains stable performance across diverse partner behaviors and coordination challenges.

\paragraph{Evaluation with Human Proxy Partner:}
We now turn our attention to evaluating all methods with a human proxy partner trained on real human data. This proxy is obtained via behavior cloning on publicly available human–human trajectories collected by Carroll et al.~\citep{carroll2019utility}. Following the same procedure, the model is trained to imitate human demonstrations and used as a fixed partner during evaluation, providing a realistic approximation of human behavior under controlled conditions. Results are reported in Table~\ref{tab:bc_partner}. Performance of all methods slightly drops compared to evaluation with the self-play population, as behavior cloning with limited human data can produce policies that favor a dominant action and occasionally stall without random perturbations, as noted in \citep{carroll2019utility}. Despite these challenges, PASD continues to achieve the highest returns, highlighting its ability to generalize effectively to previously unseen human-like partners.

\paragraph{\textcolor{black}{Human Subject Study: Real Human–AI Collaboration Evaluation}}
\textcolor{black}{To evaluate PASD in real human–AI collaboration, we conducted a controlled human-subject study following \cite{carroll2019utility}. We recruited 25 participants via Amazon Mechanical Turk (AMT), each completing two episodes of ~20 minutes: one paired with HiPT and one with PASD. The order of methods was randomized to mitigate ordering effects. To limit session duration and reduce participant fatigue, only HiPT and PASD were included, omitting other baselines.} \textcolor{black}{Participants who did not complete both episodes were excluded, leaving 19 valid participants. Trajectories and rewards were recorded for all layouts and evaluation partners. Table \ref{tab:hm_partner} reports the mean ± standard deviation of total reward across conditions. Across all layouts, PASD consistently achieved higher joint rewards with human partners than HiPT, improving human–AI collaboration by $22–47\%$, demonstrating that partner-conditioned skill discovery meaningfully enhances real-world coordination. The full experimental setup is available at our anonymized GitHub repository \footnote{\label{fn:github} https://anonymous.4open.science/r/pasd-22495/}.}

\paragraph{Qualitative Analysis of Skill Disentanglement:}
\textcolor{black}{To illustrate how PASD (blue agent) adapts to human behaviors, we conducted controlled sessions in the Cramped Room and Coordination Ring layouts. Each session lasted 80 seconds ($\sim$482 steps), simulating diverse gameplay by the human partner. In the Cramped Room, humans demonstrated varying task execution patterns (e.g., picking plates from different sides or collecting soup in distinct sequences), whereas in the Coordination Ring, humans coordinated either clockwise or counter-clockwise, requiring the AI agent to adapt its behavior to match the chosen coordination pattern.}

\textcolor{black}{Figure~\ref{fig:pasd_skill_anal0} presents cosine-similarity heatmaps computed over sequence-level embeddings, where each embedding represents a contiguous sub-trajectory during which a particular skill remains active. Figures~(\ref{fig:cramped_human_cossim}) and ~(\ref{fig:cramped_ai_cossim}) show these matrices for the Cramped Room layout (human and AI-agent embeddings, respectively), and Figures~(\ref{fig:coordination_human_cossim}) and ~(\ref{fig:coordination_ai_cossim}) show the same for the Coordination Ring layout. The heatmaps reveal strong \emph{intra-skill consistency}: similar human behavior patterns consistently trigger the same high-level skill, producing embeddings with high cosine similarity across temporally separated segments. In contrast, embeddings corresponding to different skills show low similarity, highlighting a clear \emph{inter-skill separation}. This structure reflects the effect of the contrastive intrinsic reward, which encourages embeddings generated under the same skill to remain close while pushing apart embeddings corresponding to different skills, thus enforcing temporal consistency within a skill and behavioral separation across skills. The same structure is observed in AI-agent embeddings. Once a high-level skill is selected in response to a human behavior pattern, the low-level policy produces coherent action sequences conditioned on that skill. These sequences adapt to the human behavior that triggered the skill, demonstrating that PASD aligns high-level skill selection with human intent while maintaining consistent, skill-specific action execution.}

\textcolor{black}{To further illustrate the geometric structure of the learned skill embeddings, Figure~\ref{fig:pasd_skill_umap} shows a two-dimensional UMAP projection of the same embeddings. Each point represents a skill segment, and clusters correspond to distinct skills. The visualization demonstrates that embeddings corresponding to the same skill form compact, well-separated clusters, reflecting both the temporal consistency and behavioral disentanglement enforced by the contrastive intrinsic reward.}

\textcolor{black}{To further examine the behavioral dynamics captured by PASD, Figure~\ref{fig:pasd_skill_traj} visualizes human and AI-agent trajectories in the Cramped Room layout for different skill activations. Each trajectory is represented on a 2D grid, with arrows indicating movement actions (up, down, left, right) and filled or hollow circles representing interaction and stay actions, respectively. The figure demonstrates that each high-level skill consistently maps to a distinct sub-trajectory of human behavior.In response, the AI agent selects and executes low-level actions conditioned on the activated high-level skill, producing consistent trajectories that adapt to the human's behavior pattern that triggered the skill. These results highlight that PASD not only disentangles skills at the high level but also produces predictable and skill-specific action patterns at the low level, effectively aligning agent behavior with human behavioral intent.}

\section{Conclusion}
This work presents PASD, a DHRL approach that introduces an intrinsic reward designed to enable effective human-AI coordination. The reward leverages a contrastive objective that encourages skill representations to be consistent across similar partners while remaining discriminative across diverse partner strategies. By capturing patterns shaped by partner behaviors, PASD promotes behavioral consistency and robustness, naturally mitigating shortcut learning that can arise from spurious information in the state space. Our experiments in Overcooked-AI demonstrate that PASD learns transferable skills that generalize across a wide range of partners, providing a foundation for more adaptive and efficient collaborative agents.

\appendix

\section{ Layout Challenges and the Need for Adaptive Skill Learning}
\label{appendix:layouts}

Each Overcooked-AI layout presents unique coordination challenges requiring agents to adapt to diverse partners with varying skill levels and play styles. In \textit{Cramped Room}, collisions are frequent due to limited space, necessitating adaptable turn-taking and collision avoidance. \textit{Asymmetric Advantages} features partners specializing in different roles, requiring flexible skill activation for complementary behavior. \textit{Coordination Ring} enforces a looped workflow, demanding alignment with partners’ directional preferences. In \textit{Counter Circuit}, interactions occur via counters, making timing and item exchange strategies critical. \textit{Forced Coordination} imposes physical separation, emphasizing sequenced cooperation and dynamic routines.

In addition to the environment reward of $+20$ for each successful soup delivery, we incorporate shaped rewards to facilitate effective coordination with diverse partners. Picking up or placing an onion into a pot yields a small positive reward of $+3$, while a penalty of $-20$ is applied when the partner delivers a soup. These rewards are used in all layouts except \emph{Forced Coordination}, where strict role asymmetry naturally enforces task specialization. By providing intermediate feedback, shaped rewards guide the agent to engage in complementary sub-tasks and adapt its behavior to align with the actions and strategies of different partners, promoting robust collaboration across all layouts.

\section{Implementation Details}
\label{appendix:implementation}

We use consistent training settings across all layouts for the PPO objective. Specifically, the entropy loss coefficient is set to $0.01$ for both high- and low-level policies and linearly decays to zero over the course of training. The value function coefficient is fixed at $0.5$, and the PPO clipping coefficient is set to $0.05$. A complete summary of general hyperparameters is provided in Table~\ref{tab:general_hparams}.  

Certain training parameters, such as the initial learning rate and decay schedule, are tailored to each layout to account for differing coordination challenges. Table~\ref{tab:layout_hparams} summarizes these layout-specific settings.   

\begin{table}
\centering
\caption{Hyperparameters applied across all layouts.}
\label{tab:general_hparams}
\begin{tabular}{lc}
\toprule
\textbf{Hyperparameter} & \textbf{Value} \\
\midrule
Entropy coefficient & $0.01 \rightarrow 0$ (linear decay) \\
Value function coefficient & $0.5$ \\
Clipping coefficient & $0.05$ \\
Optimizer & Adam \\
Discount factor $\gamma$ & $0.99$ \\
GAE parameter  & $0.98$ \\
Batch size & $64$ per environment \\
Parallel environments & $30$ \\
\bottomrule
\end{tabular}
\end{table}

\begin{table}
\centering
\caption{Layout-specific training parameters. The learning rate decays linearly from the initial value to the initial value divided by the decay ratio over training.}
\label{tab:layout_hparams}
\begin{tabular}{lcc}
\toprule
\textbf{Layout} & \textbf{Initial Learning Rate} & \textbf{Decay Ratio} \\
\midrule
Cramped Room            & $1.0 \times 10^{-3}$ & $3$ \\
Asymmetric Advantages   & $1.0 \times 10^{-3}$ & $3$ \\
Coordination Ring       & $6.0 \times 10^{-4}$ & $1.5$ \\
Forced Coordination     & $8.0 \times 10^{-4}$ & $2$ \\
Counter Circuit         & $8.0 \times 10^{-4}$ & $3$ \\
\bottomrule
\end{tabular}
\end{table}

\printcredits

\section*{Declaration of competing interest}
The authors declare that they have no known competing financial interests or personal relationships that could have appeared to influence the work reported in this paper.
\section*{Funding}
This research was funded by the Asian Office of Aerospace Research and Development through Grant 23IOA087.
\section{Declaration of generative AI use}
Generative AI tools were used only for language polishing and grammar improvement during the preparation of this manuscript. The authors reviewed and approved the final manuscript.













\bibliographystyle{unsrt}
\bibliography{cas-refs}



\end{document}